\documentclass[english]{article}
\usepackage{mathptmx}
\usepackage{helvet}
\usepackage{courier}
\usepackage[T1]{fontenc}
\usepackage[utf8]{inputenc}
\usepackage{color}
\usepackage{babel}
\usepackage{float}
\usepackage{textcomp}
\usepackage{amsmath}
\usepackage{amsthm}
\usepackage{amssymb}
\usepackage{graphicx}

\makeatletter

\newcommand{\noun}[1]{\textsc{#1}}
\providecommand{\tabularnewline}{\\}

\numberwithin{equation}{section}
\numberwithin{figure}{section}
\newcommand{\lyxaddress}[1]{
\par {\raggedright #1
\vspace{1.4em}
\noindent\par}
}
 \theoremstyle{definition}
 \newtheorem*{defn*}{\protect\definitionname}
  \theoremstyle{remark}
  \newtheorem*{claim*}{\protect\claimname}

\usepackage{url} 

  \definecolor{docnotelinkcolor}{rgb}{0,0,0.4}

\makeatother

  \providecommand{\claimname}{Claim}
  \providecommand{\definitionname}{Definition}

\begin{document}

\title{Sheaves: A Topological Approach to Big Data}

\author{Linas Vepštas\textsuperscript{{*}}}

\date{12 November 2017}
\maketitle
\begin{abstract}
This document develops general concepts useful for extracting knowledge
embedded in large graphs or datasets that have pair-wise relationships,
such as cause-effect-type relations. Almost no underlying assumptions
are made, other than that the data can be presented in terms of pair-wise
relationships between objects/events. This assumption is used to mine
for patterns in the dataset, defining a reduced graph or dataset that
boils-down or concentrates information into a more compact form. The
resulting extracted structure or set of patterns are manifestly symbolic
in nature, as they capture and encode the graph structure of the dataset
in terms of a (generative) grammar. This structure is identified as
having the formal mathematical structure of a sheaf. In essence, this
paper introduces the basic concepts of sheaf theory into the domain
of graphical datasets.

DRAFT: This is an unfinished draft; the last 1/4th of the document
needs a complete make-over.
\end{abstract}

\lyxaddress{\textsuperscript{{*}}Hanson Robotics; SingularityNET; <linasvepstas@gmail.com>}

\lyxaddress{ACM Subject Classification:\\
• Theory of computation—Models of computation;500,\\
• Theory of computation—Formal languages and automata theory—Grammars
and context-free languages;500,\\
• Theory of computation—Design and analysis of algorithms—Graph algorithms
analysis;500}

\section*{Intro}

This document presents some definitions and vocabulary for working
with datasets that contain complex relationships, applicable to a
large variety of application domains. The concepts borrow from graph
theory, and several other areas of mathematics. The goal is to define
a way of thinking about complex graphs, and how they can be simplified
and condensed into simpler graphs that ``concentrate'' embedded
knowledge into a more manageable size. The output of the process is
a grammar that summarizes or captures the significant or important
relationships.

The ideas described here are not terribly complex; they represent
a kind-of ``folk knowledge'' generally known to a number of practitioners.
However, I am not currently aware of any kind of presentation of this
information, either in review/summary form, or as a fully articulated
book or text. The background knowledge appears to be scattered across
wide domains, and occur primarily in highly abstract settings, outside
of the mainstream computer-science and data-analysis domain. Thus,
this document tries to provide an introduction to these concepts in
a plain-spoken language. The hope is to be precise enough that there
will be few complaints from the mathematically rigorous-minded, yet
simple enough that ``anyone'' can follow through and understand.

Some examples are provided, primarily drawn from linguistics. However,
the concepts are generally applicable, and should prove useful for
analyzing any kind of dataset expressed with pair-wise relationships,
but containing hidden (non-obvious) complex cause-and-effect relationships.
Such datasets include genomic and proteomic data, social-graph data,
and even such social policy information.

Consider the example of determining the effectiveness of educational
curricula. When teaching students, one never teaches advanced topics
until foundations are laid. Yet many students struggle. Given raw
data on a large sample of students, and the curricula they were subjected
to, can one discern sequences and dependencies of cause-and-effect
in this data? Can one find the most effective curriculum to teach,
that advances the greatest number of students? Can one discover different
classes of students, some who respond better to one style than another?
My belief is that these questions can not only be answered, but that
the framework described here can be used to uncover this structure.

Another example might be the analysis of motives and actions in humans.
This includes analysis from real life, as well as the narratives of
books and movies. In a book setting, the author cannot easily put
characters into action until some basic sketch of personality and
motives is developed. Motives can't be understood until a setting
is established. If one can break down a large number of books/movies
into pairs of related facts/scenes/remarks/actions, one can then extract
a grammar of relationships, to see exactly what is involved in the
movement of a narrative from here to there.

Much of this document is devoted to stating definitions for a few
key structures used to talk about the general problem of discerning
relationships and structure. The definitions are inspired by and draw
upon concepts from algebraic topology, but mostly avoid both the rigor
and the difficulty of that topic. 

The definitions provide a framework, rather than an algorithm. It
is up to the user to provide some mechanism for judging similarity
- and this can be anything: some neural net, Bayesian net, Markov
chain, or some vector space or SVM-style technique; the overall framework
is agnostic as to these details. The goal is to provide a way of talking
about, thinking about and presenting data so that the important knowledge
contained in it is captured and described, boiled down to a manageable,
workable state from a large raw dump of pair-relationship data.

Currently, the ideas described here are employed in a machine-learning
project that attempts to extract the structure of natural language
in an unsupervised way. Thus, the primary, detailed examples will
come from the natural language domain. The theory should be far more
general than that.

This document resides in, accompanies source code that implements
the ideas here. Specifically, it is in \href{https://github.com/opencog/atomspace/tree/master/opencog/sheaf}{https://github.com/opencog/atomspace/tree/master/opencog/sheaf}
and it spills over into other files, such as \href{https://github.com/opencog/opencog/blob/master/opencog/nlp/learn/scm/gram-class.scm}{https://github.com/opencog/opencog/blob/master/ opencog/nlp/learn/scm/gram-class.scm}
This code is in active development, and is likely to have changed
by a lot since this was written. This document is \emph{not} intended
to describe the code; rather, it is meant to describe the general
underlying concepts.

For the mathematically inclined, please be aware that the concepts
described here touch on the tiniest tips of some very deep mathematical
icebergs, specifically in parsing, type theory and category theory.
I have no hope of providing the needed background, as these fields
are sophisticated and immense. The reader is encouraged to study these
on their own, especially as they are applied in computer science and
linguistics. There are many good texts on these topics.

This document is organized as follows. The first part of provides
a definition of a ``section'' of a graph. A section is a lot like
a subgraph, except that it explicitly indicates which edges were cut
to form the subgraph. The next part defines and articulates the concept
of projection, and shows how it can be used to form quotients. The
quotients or projections are termed ``stalks'', and, because each
stalk comes festooned with connectors, they can be thought to resemble
corn-stalks. The next part shows how stalks can be tied together to
form sheaves, and reviews the axioms of sheaf theory to show that
this name is appropriate.

After this comes a lighting review of how data mining, pattern mining
and clustering can be viewed in the context of sheaves. After this
come two asides: a quick sketch of type theory, illustrating the interplay
between data-mined patterns and the concept of types. Another aside
reviews the nature of parsing, illustrating that parsing algorithms
implement the gluing axiom of sheaves, viz, that gluing and parsing
are the same thing. The final part examines polymorphic behavior.
Polymoprhism is that point where syntax begins to touch semantics,
where deep structure becomes distinguished from surface structure.

\section*{Sections}

Begin with the standard definition of a graph.
\begin{defn*}
A \noun{graph} $G=\left(V,E\right)$ is an ordered pair $\left(V,E\right)$
of two sets, the first being the set $V$ of vertices, and the second
being the set $E$ of edges. An edge $e\in E$ is a pair $\left(v_{1},v_{2}\right)$
of vertices, where every $v_{k}$ \emph{must} be a member of $V$.
That is, edges in $E$ can only connect vertexes in $V$, and not
to something else. $\diamond$
\end{defn*}
For directed graphs, the vertex ordering in the edge matters. For
undirected graphs, it does not. The subsequent will mostly leave this
distinction unspecified, and allow either (or both) directed and undirected
edges, as the occasion and the need fits. Distinguishing between directed
and undirected graphs is not important, at this point. In most of
what follows, it will usually be assumed that there are no edges with
$v_{1}=v_{2}$ (loops that connect back to themselves) and that there
is at most one edge connecting any given pair of vertexes. These assumptions
are being made to simplify the discussion; they are not meant to be
a fundamental limitation. It just makes things easier to talk about
and less cluttered at the start. The primary application does not
require either construct, and it is straight-forward to add extensions
to provide these features. Similar remarks apply to graphs with labeled
vertexes or edges (such as ``colored'' edges, vertexes or edges
with numerical weights on them, \emph{etc}). Just keep in mind that
such additional markup may appear out of thin air, later on.

Besides the above definition, there are other ways of defining and
specifying graphs. The one that will be of primary interest here will
be one that defines graphs as a collection of sections. These, in
turn, are composed of seeds.
\begin{defn*}
A \noun{seed} is a vertex and the set of edges that connect to it.
That is, it is the pair $\left(v,E_{v}\right)$ where $v$ is a single
vertex, and $E_{v}$ is a set of edges containing that vertex, i.e.
that set of edges having $v$ as one or the other endpoint. The vertex
$v$ may be called the \noun{germ} of the seed. For each edge in the
edge set, the other vertex is called the \noun{connector}.$\diamond$
\end{defn*}
It should be clear that, given a graph $G$, one can equivalently
describe it as a set of seeds (one simply lists all of the vertexes,
and all of the edges attached to each vertex). The converse is not
``naturally'' true. Consider a single seed, consisting of one vertex
$v_{1}$, and a single edge $e=\left(v_{1},v_{2}\right)$. Then the
pair $\left(V,E\right)$ with $V=\left\{ v_{1}\right\} $ and $E=\left\{ \left(v_{1},v_{2}\right)\right\} $
is \emph{not} a graph, because $v_{2}$ is missing from the set $V$.
Of course, we could implicitly include $v_{2}$ in the collection
of vertexes, but this is not ``natural'', if one is taking the germs
of the seeds to define the vertexes of the graph.

Thus, given a seed, each edge in that seed has one ``connected''
endpoint, and one ``unconnected'' endpoint. The ``connected''
endpoint is that endpoint that is $v$. The other endpoint will commonly
be called the \noun{connector}; equivalently, the edge can be taken
to be the connector. Perhaps it should be called a half-edge, as one
end-point is specified, but missing.

The seed can be visualized as a ball, with a bunch of sticks sticking
out of it. A burr one might collect on one's clothing. One can envision
a seed as an analog of an open set in topology: the center (the germ)
is part of the set, and then there's some more, but the boundary is
not part of the set. The vertexes on the unconnected ends of the edges
are not a part of the seed.
\begin{figure}[h]
\caption{A seed}

\includegraphics[width=0.25\columnwidth]{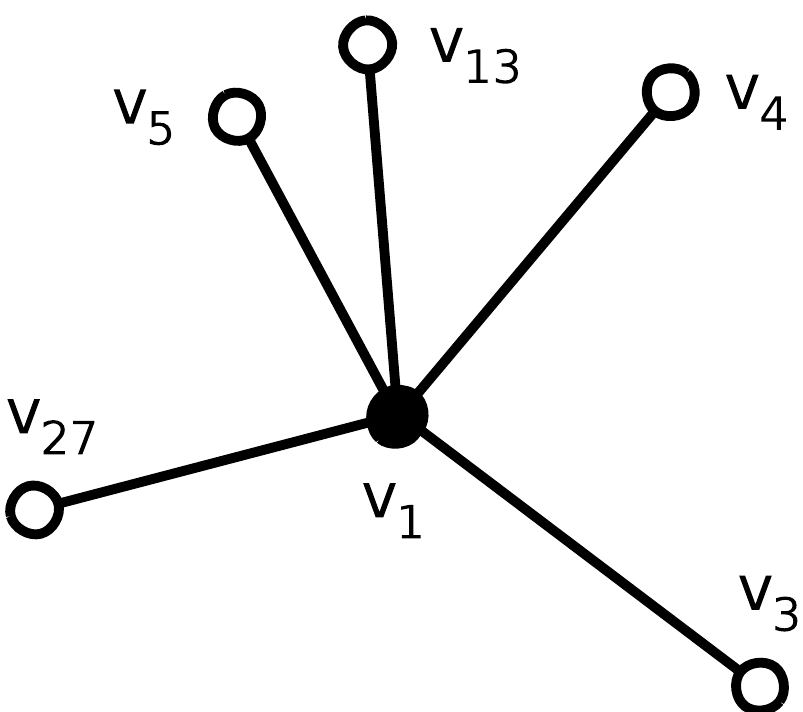}

\end{figure}

Just as one can cover a topological space with a collection of open
sets, so one can also cover a graph with seeds. This analogy is firm:
if one has open sets $U_{i}$ and $U_{j}$ and $U_{i}\cap U_{j}\ne\emptyset$
then one can take $U_{i}$ and $U_{j}$ to be vertices, and $U_{i}\cap U_{j}$
to be an edge running between them.

More definitions are needed to advance the ideas of connecting and
covering.
\begin{defn*}
A \noun{section} is a set of seeds. $\diamond$
\end{defn*}
It should be clear that a graph $G$ can be expressed as section;
that section has the nice property that all of the germs appear once
(and only once) in the set $V$ of $G$, and that all of the edges
in $E$ appear twice, once each in two distinct seeds. This connectivity
property motivates the following definition: 
\begin{defn*}
Given a section $S$, a \noun{link} is any edge $\left(v_{1},v_{2}\right)$
where both $v_{1}$ and $v_{2}$ appear as germs of seeds in $S$.
Two seeds are \noun{connected} when there is a link between them.
$\diamond$
\end{defn*}
This definition of a link is imprecise. A more proper, technical definition
is that a link can be formed only when the germ $v_{1}$ has $v_{2}$
as a connector, and also, at the same time, the germ $v_{2}$ has
$v_{1}$ as a connector; only then can the two be joined together.
The joining is meant to be optional, not mandatory: just because a
section contains connectors that can be joined, it does not imply
that they must be. The joining is also meant to consume the connectors
as a resource: once two connectors have been connected, neither one
is free to make connections elsewhere.

\begin{figure}[h]
\caption{Two linked (connected) seeds}
\includegraphics[width=0.35\columnwidth]{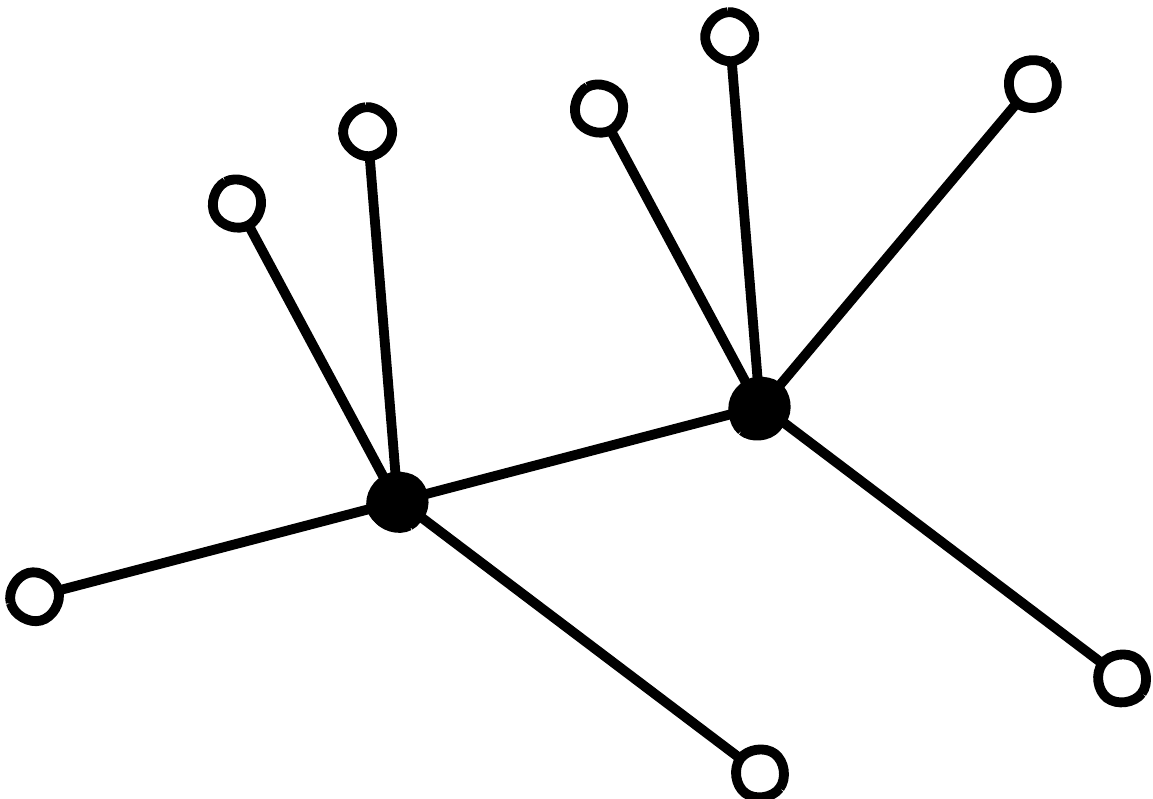}

\end{figure}
The use of links allows the concepts of paths and connectivity, taken
from graph theory, to be imported into the current context. Thus,
one can obviously define:
\begin{defn*}
A \noun{connected section}, or a \noun{contiguous section} is a section
where every germ is connected to every other germ via a path through
the edges. $\diamond$
\end{defn*}
In graph theory, this would normally be called a ``connected graph'',
but we cannot fairly call it that because the seeds and sections were
defined in such a way that they are not graphs; they only become graphs
when they are fully connected. Never-the-less, it is fairly safe and
straight-forward to apply common concepts from graph-theory. Sections
are almost like graphs, but not quite.

Note that there are two types of edges in a section: those edges that
connect to nothing, and those edges that connect to other seeds in
that section. Henceforth, the unconnected edges will be called connectors
(as defined above), while the fully-connected edges will be called
links (also defined above). Connectors can be thought of as a kind-of
half-edge: incomplete, missing the far end, while links are fully
connected, whole.

Seeds and sections can (and should!) be visualized as hedgehogs -
a body with spines sticking out of it - the connectors can be thought
of as the spiny bits sticking out, waiting to make a connection, while
the hedgehog body is that collection of vertices and the fully-connected
links between them.
\begin{figure}[h]
\caption{A connected section}
\includegraphics[width=0.25\columnwidth]{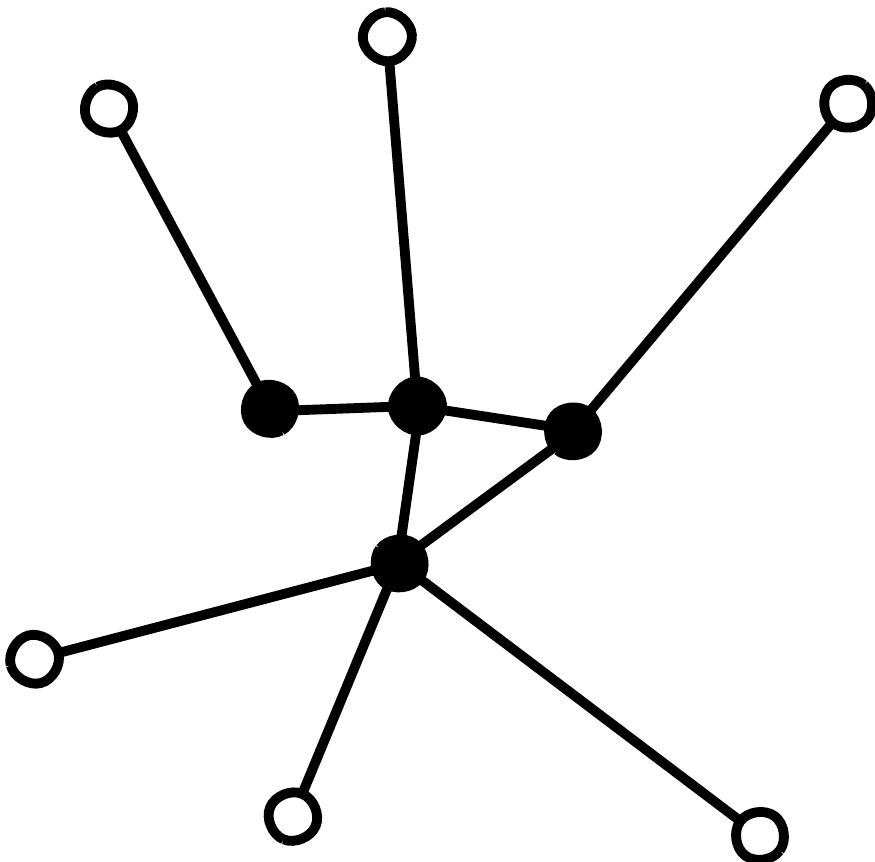}

\end{figure}

Implicit in the above definitions was that, during link formation,
an edge is only allowed to connect to another seed if and only if
the connector matches the germ. That is, if $\left(v_{1},v_{2}\right)$
is an edge rooted in the seed for $v_{1}$ and if $\left(v_{3},v_{4}\right)$
is an edge rooted in the seed for $v_{3}$, then these two can form
a link if and only if $v_{2}=v_{3}$ and $v_{4}=v_{1}$. That is,
the connectors are typed: they can only connect to seeds that are
of the same type as the unconnected end of the edge.

This motivates a different way of looking at seeds: they can be visualized
as jigsaw puzzle pieces, where any given tab on one jigsaw piece can
fit into one and only one slot on another jigsaw piece. This union
of a tab+slot is the link. Connectors must be of the same type in
order to be connectible. The types of the connectors will later be
seen to be the same thing as the types of type theory; that is, they
are bona-fide types, in the proper sense of the word. 

\begin{figure}[h]
\caption{Joining two connectors to form a link}

\includegraphics[width=0.4\columnwidth]{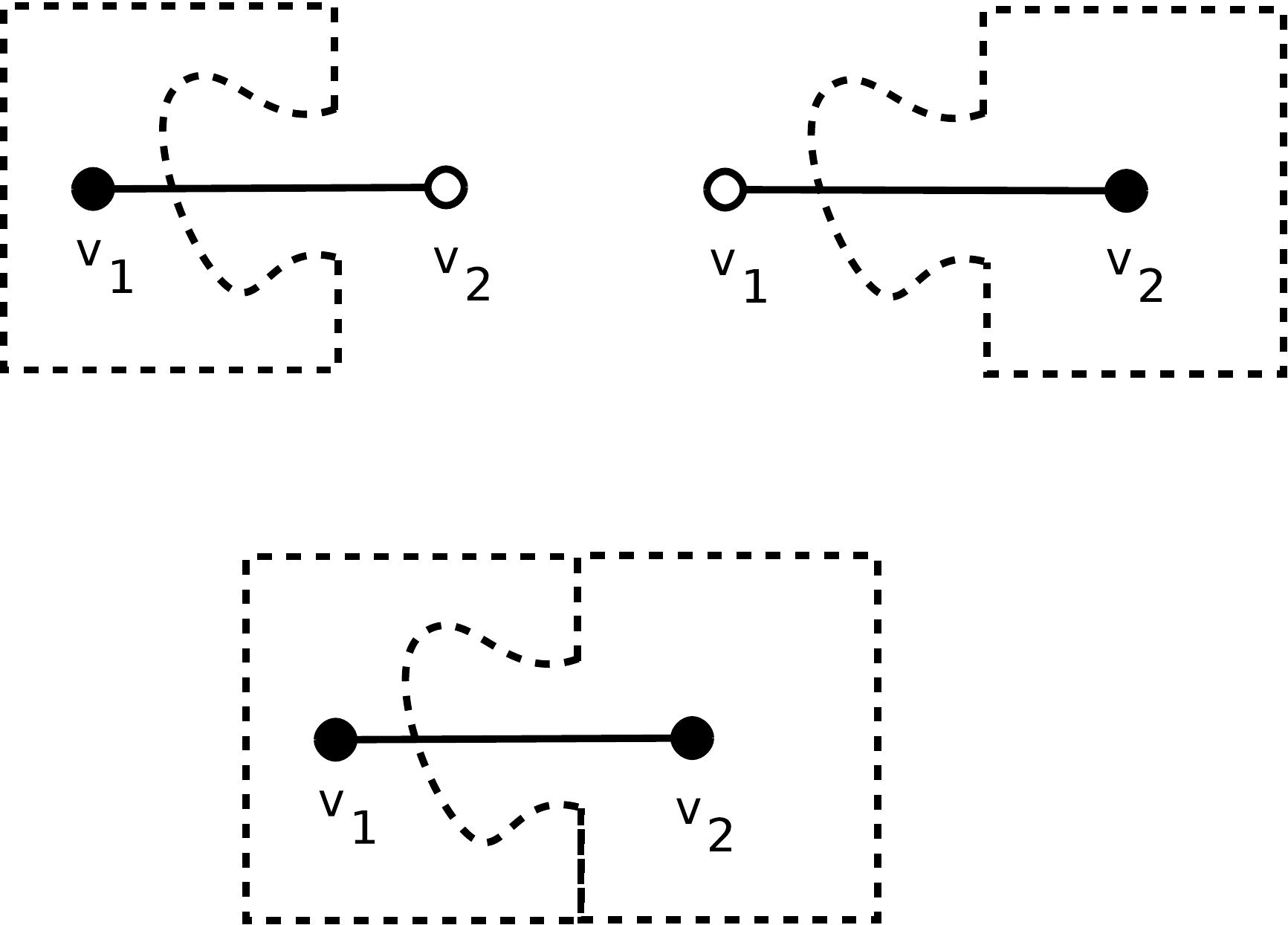}

\end{figure}

The jigsaw puzzle-piece illustration is not uncommon in the literature;
such illustrations are explicitly depicted in a variety of settings.\cite{Sleator1991,Coecke2010,Kart2014,Baez2009}
The point being illustrated here is that the connectors need not be
specific vertexes, they can be vertex types, where any connector of
the appropriate type is allowed to connect. This can be formalized
in an expanded definition of a seed. A provisional definition of a
type is needed, first.
\begin{defn*}
A \noun{type} is a set of vertexes. Notationally, $t=\left\{ v_{a},v_{b},\cdots\right\} $.
$\diamond$
\end{defn*}
This allows the jigsaw concept to be expressed more formally. 
\begin{defn*}
A \noun{seed} is a vertex and the set of connector types that connect
to it. That is, it is the pair $\left(v,C_{v}\right)$ where $v$
is a vertex, and $C_{v}$ is a set of connector types containing that
vertex, i.e. that set of edges having $v$ as one endpoint and a type
as the other endpoint. That is, $C_{v}=\left\{ \left(v,t_{a}\right),\left(v,t_{b}\right),\cdots\right\} $.
A single pair $\left(v,t\right)$ can be called a \noun{connector
type}. $\diamond$
\end{defn*}
The capital letter $C$ is used to remind one that members of the
set are connectors. The intent of specifying connector types is exactly
what the jigsaw-puzzle paradigm suggests: links can be created, as
long as the types match up. This is formalized by expanding the definition
of a link. 
\begin{defn*}
Given a section $S$, a \noun{link} between seeds $s_{1}=\left(v_{1},C_{1}\right)$
and $s_{2}=\left(v_{2},C_{2}\right)$ is any edge $\left(v_{1},v_{2}\right)$
where $v_{1}$ is in one of the types in $C_{2}$ and $v_{2}$ is
in one of the types in $C_{1}$. That is, there exists a pair $\left(v_{1},t_{a}\right)\in C_{1}$
such that $v_{2}\in t_{a}$ and, symmetrically, there exists a pair
$\left(v_{2},t_{b}\right)\in C_{2}$ such that $v_{1}\in t_{b}$ .
Two seeds are \noun{connected} when there is a link between them.
$\diamond$
\end{defn*}
As before, the creation of links is meant to be optional, not forced.
As before, the connectors are meant to be consumable: once connected,
they cannot be used again. The figure below illustrates the idea.

\begin{figure}[h]
\caption{Seed connectors might be types, not vertexes}

\includegraphics[width=0.35\columnwidth]{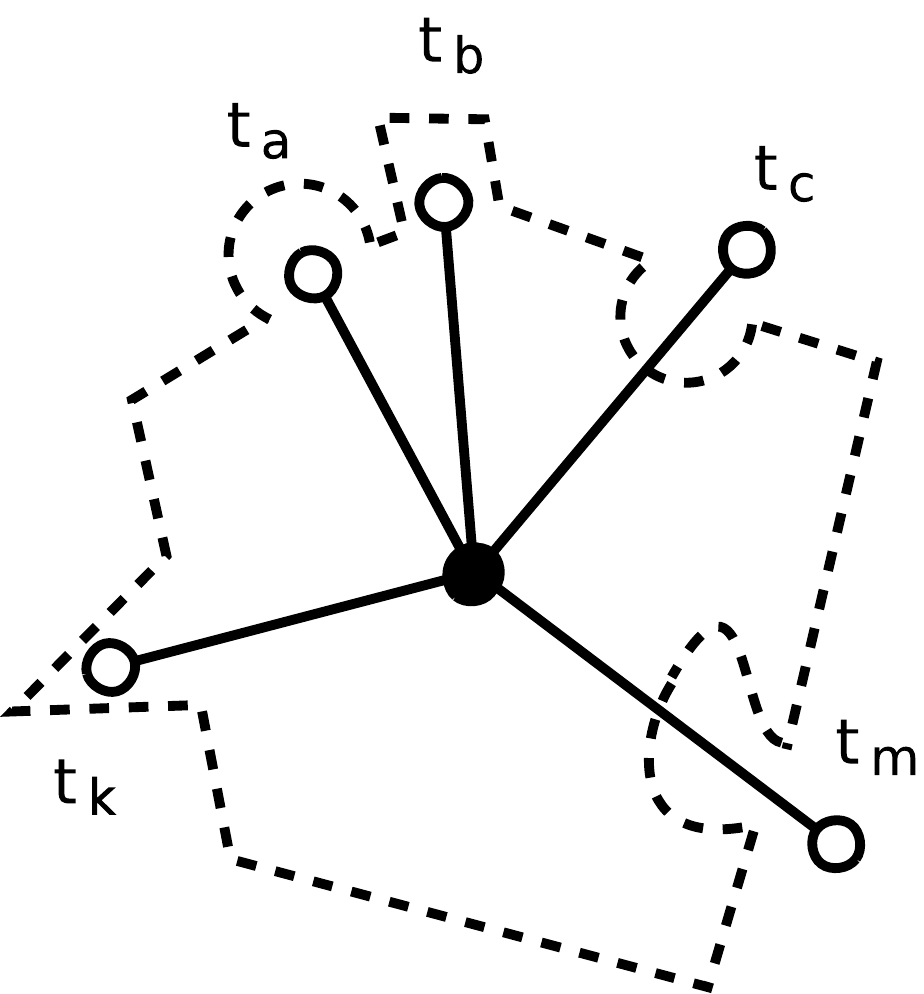}
\end{figure}

Its important to realize that the standard approach to graph theory
has been left behind. Although it is possible to hook up seeds to
form a graph, it is also possible to have a collection of seeds that
is not a graph: the category of sections contain the category of graphs
as a subset. Extending the notion of a connector to be the notion
of a connector-type in particular plays considerable violence to the
notion of graph theory. As long as the narrower definition of seed
was used, one could imagine that a collection of seeds could be assembled
into a graph, and that assembly is unique. Once connector types are
introduced, the possibility that there are multiple, non-unique assemblages
of seeds becomes possible. A graph can be disassembled into seeds,
and, if one is careful to label vertexes and edges in a unique way,
that collection can be viewed as isomorphic to the original graph.
If one is not careful, sloppily assigning labels or avoiding them
entirely, the collection can have multiple non-isomorphic re-assemblies.
The ability to be sloppy in this way is one of the appeals, one of
the benefits of working with seeds and sections. They provide ``elbow
room'' not available in (naive) graph theory.

\subsection*{Why sections?}

Whats the point of introducing this seemingly non-standard approach
to something that looks a lot like graph theory? There are several
reasons.
\begin{itemize}
\item From a computational viewpoint, sections have nice properties that
a list of vertexes and edges do not. Given a single seed, one ``instantly''
knows \emph{all} of the edges attached to its germ: they are listed
right there. By contrast, given only a graph description, one has
to search the entire list $E$ for any edges that might contain the
given vertex. Computationally, searching large lists is inefficient,
especially so for very large graphs.
\item The subset of a section is always a section. This is not the case
for a graph: given $G=\left(V,E\right)$, some arbitrary subset of
$V$ and some arbitrary subset of $E$ do not generally form a graph;
one has to apply consistency conditions to get a subgraph.
\item A connected section behaves very much like a seed: just as two seeds
can be linked together to form a connected section, so also two connected
sections can be linked together to form a larger connected section.
Both have a body, with spines sticking out. The building blocks (seeds),
and the things built from them (sections) have the same properties,
lie in the same class. Thus, one has a system that is naturally ``scalable'',
and allows notions of similarity and scale invariance to be explored.
There is no need to introduce additional concepts and constructions.
\item Given two seeds, one can always either join them (because they connect)
or it is impossible to connect them. Either way, one knows immediately.
Graphs, in general, cannot be joined, unless one specifies a subgraph
in each that matches up. Locating subgraphs in a graph is computationally
expensive; verifying subgraph isomorphism is computationally expensive.
\item The analogy between graphs and topology, specifically between open
sets and seeds and the intersection of open sets and edges, allows
concepts and tools to be borrowed from algebraic topology.
\end{itemize}
If we stop here, not much is accomplished, other than to define a
somewhat idiosyncratic view of graph theory. But that is not the case;
the concept of seeds and sections are needed to pursue more complex
constructions. They provide a tool to study natural language and other
systems.

\subsection*{Example: Biochemical reaction type}

An example of a seed applied to the biochemical domain would be the
phosphorylation of ADP to ATP, shown in the figure below.
\begin{center}
\includegraphics[width=0.35\columnwidth]{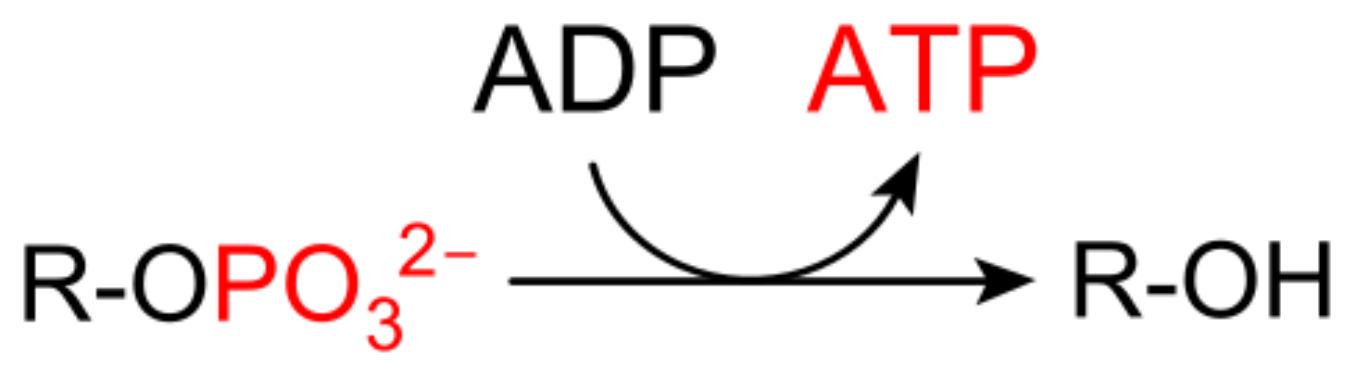}
\par\end{center}

The germ of the seed is the point where the semi-circle kisses the
line: not labeled here, the germ would be succinate-CoA ligase. The
connectors are labeled with their types, and the arrows provide directionality.
The connector types clearly indicate what can be linked to what: this
particular seed, when linked, \emph{must} link to a source of ADP,
or a source of phosphate, or a sink if ATP or a sink of hydroxyls,
if it is to be validly linked into any part of a connected section.
In this example, ADP and ATP can both be treated as simple connectors,
while R-OH does name a type: R can be any moiety. Implicit here, but
not explicit in the seed, is that the R group on both connectors must
be the same.

An example of a connected section would be the Krebs cycle, taken
as a whole:
\begin{center}
\includegraphics[width=0.55\columnwidth]{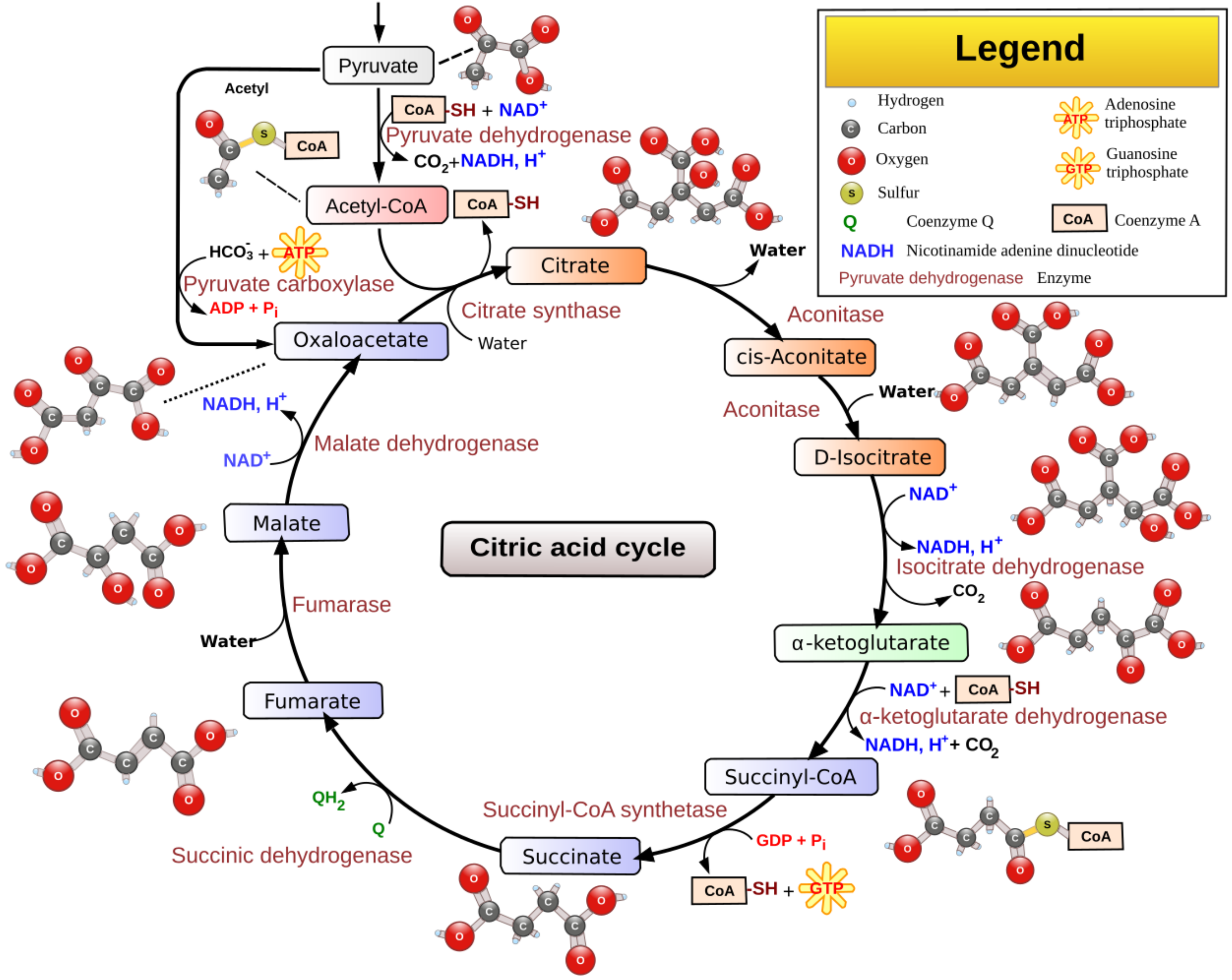}
\par\end{center}

Each distinct reaction constitutes a seed; the heavy lines forming
the cycle are the links internal to the section, and each tangent
arrow is a pair of connectors, with one end of the arrow being an
unconnected reaction input, and the other end of the arrow an unconnected
reaction product. Thus, for example, connector types include NAD,
NADH, water and ATP, among others. These connectors are free to be
attached to other seeds or sections.

This example may seem dubious, at this point of the presentation.
That it is a valid example should become clear with further development
of the general principles in what follows.

\subsection*{Similar concept: Link Grammar}

Readers familiar with Link Grammar\cite{Sleator1991,Sleator1993}
should have recognized seeds as being more or less the same thing
as ``disjuncts'' in Link Grammar. The formal definition for Link
Grammar disjuncts are a bit more complicated than seeds, and is expanded
on in later sections. To lay that groundwork, however, consider an
unlabeled dependency parse for the sentence ``this is an example'',
shown in the figure below.

\begin{figure}[h]
\caption{A dependency parse decomposed into four seeds}
\includegraphics[width=0.85\columnwidth]{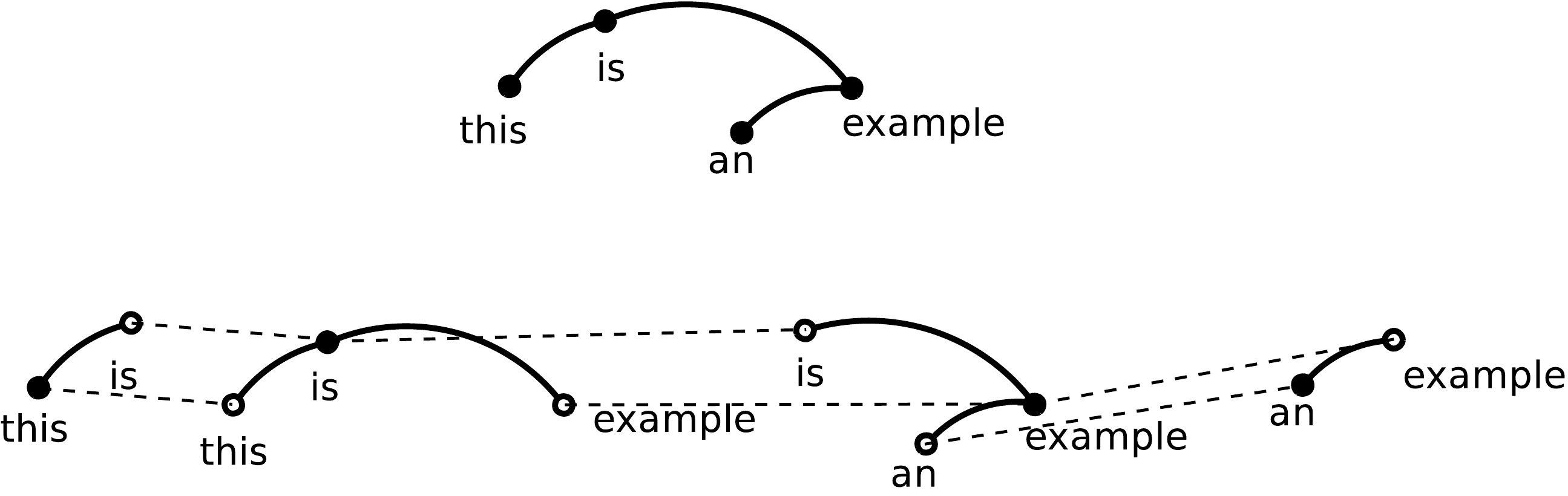}

\end{figure}

The dependency parse is shown as a graph, with four vertexes. Below,
the parse is decomposed into the component seeds; as always, the open
dots are connectors, the closed dots are the germs. Using the notation
$\left(v,C_{v}\right)$ for a seed, where $C_{v}=\left\{ \left(v,v_{a}\right),\left(v,v_{b}\right),\cdots\right\} $,
these seeds can be textually written as\\

\qquad{}%
\begin{minipage}[t]{0.8\columnwidth}%
\textsf{this: \{(this, is+)\}}

\textsf{is: \{(is, this-), (is, example+)\}}

\textsf{an: \{(an, example+)\}}

\textsf{example: \{(example, is-), (example, an-)\}}%
\end{minipage}\\
\\
\\
The above vertex: edge-list notation is a bit awkward and hard to
read. A simpler notation conveying the same idea is\\

\qquad{}%
\begin{minipage}[t]{0.8\columnwidth}%
\textsf{this: is+;}

\textsf{is: this- \& example+;}

\textsf{an: example+;}

\textsf{example: an- \& is-;}%
\end{minipage}\\
\\
\\
In both textual representations, the pluses and minuses are used to
indicate word-order: minuses to the left, pluses to the right. This
is an additional decoration added to the connectors, needed to indicate
and preserve word-order, but not a part of the core definition of
a seed. The ampersand is not symmetric, but enforces order; this is
not apparent here, but is required for the proper definition.

In Link Grammar, the objects to the right of the colon are called
``disjuncts''. The name comes from the idea that they disjoin colocational
extractions. After observing a large corpus, one might find that\\

\qquad{}%
\begin{minipage}[t]{0.8\columnwidth}%
\textsf{is: (this- \& example+) or (banana- \& fruit+) or (apple-
\& green+);}%
\end{minipage}\\
\\
which indicates that sentences such as ``a banana is a kind of fruit''
or ``this apple is green'' were observed and parsed into (unlabeled)
dependencies.

\subsection*{Similar concept: lambda notation}

Linguistics literature sometimes describes similar concepts using
a lambda-calculus notation. For example, one can sort-of envision
the expression $\lambda M.xyz$ as a seed with the germ $M$ and with
connectors $x$, $y$ and $z$. This notation has been used to express
the concept of a seed, as described above. For example, Poon and Domingos\cite{Poon2009}
write $\lambda y\lambda x.\mbox{borders}(x,y)$ to represent the attachments
of the word ``borders'' as a synonym for ``is next to''. This
is illustrated with the verb-phrase $\lambda y\lambda x.\mbox{borders}(x,y)(\mbox{Idaho})$
which beta-reduces to the verb-phrase $\lambda x.\mbox{borders}(x,\mbox{Idaho})$
to indicate that $x$ is next to Idaho. The utility of this device
becomes apparent because one can use this same notation to write $\lambda y\lambda x.\mbox{is\_next\_to}(x,y)$
and $\lambda y\lambda x.\mbox{shares\_a\_border\_with}(x,y)$ as synonymous
phrases. The lambda notation allows $x$ and $y$ to be exposed as
connectors, while at the same time hiding the links that were required
to assemble seeds for ``next'', ``is'', and ``to'' into a phrase.
That is, $\lambda y\lambda x.\mbox{is\_next\_to}(x,y)$ is an example
of a connected section, having $x$ and $y$ as the externally exposed
connectors and the internal links between ``next'', ``is'', and
``to'' hidden.

The problem with this notation is that, properly speaking, lambda
calculus is a system for generating and working with strings, not
with graphs, and lambdas are designed to perform substitution (beta-reduction),
and not for connecting things. 

That is, lambda terms are always strings of symbols, and the variables
bound by the lambda are used to perform substitutions. To illustrate
the issue, suppose that $M$ above is $axbyczd$ and suppose that
$\lambda N.w=ewf$. Can these be ``connected'' together, linked
together like seeds? No: if one tried to ``connect'' $N$ to $z,$
one has the beta-reduction $(\lambda M.xyz)\lambda N.w\rightarrow\lambda axbycewfd.xyw$.
There is no way to express some symmetric version of this, because
$(\lambda N.w)\lambda M.xyz\rightarrow\lambda eaxbyczdf.xyz$ which
is hardly the same. Now, of course, lambda calculus has great expressive
power, and one could invent a way encoding graph theory, and/or seeds,
in lambda calculus; however, doing so would result in verbose and
complex system. Its easier to work with graphs directly, and just
sleep peacefully with the knowledge that one could encode them with
lambdas, if that is what your life depended on.

Note also that there have been extensions of the ideas of lambda calculus
to graphs; however, those extensions cling to the fundamental concept
of beta reduction. Thus, one works with graphs that have variables
in them. Given a variable, one plugs in a graph in the place of that
variable. The OpenCog \href{http://wiki.opencog.org/w/PutLink}{PutLink}
works in exactly this way. The beta-reduction is fundamentally not
symmetrical: putting A into B is not the same as putting B into A.
The concept of ``connecting'' in a symmetric way doesn't arise.

\subsection*{Similar concept: tensor algebra}

The \href{https://en.wikipedia.org/w/Tensor_algebra}{tensor algebra}
is an important mathematical construct underlying large parts of mathematical
analysis, including the theory of vector spaces, the theory of Hilbert
spaces, and, in physics, the theory of quantum mechanics. 

\begin{figure}[h]
\caption{A tensor with three input wires and two output wires}

\includegraphics[width=0.3\columnwidth]{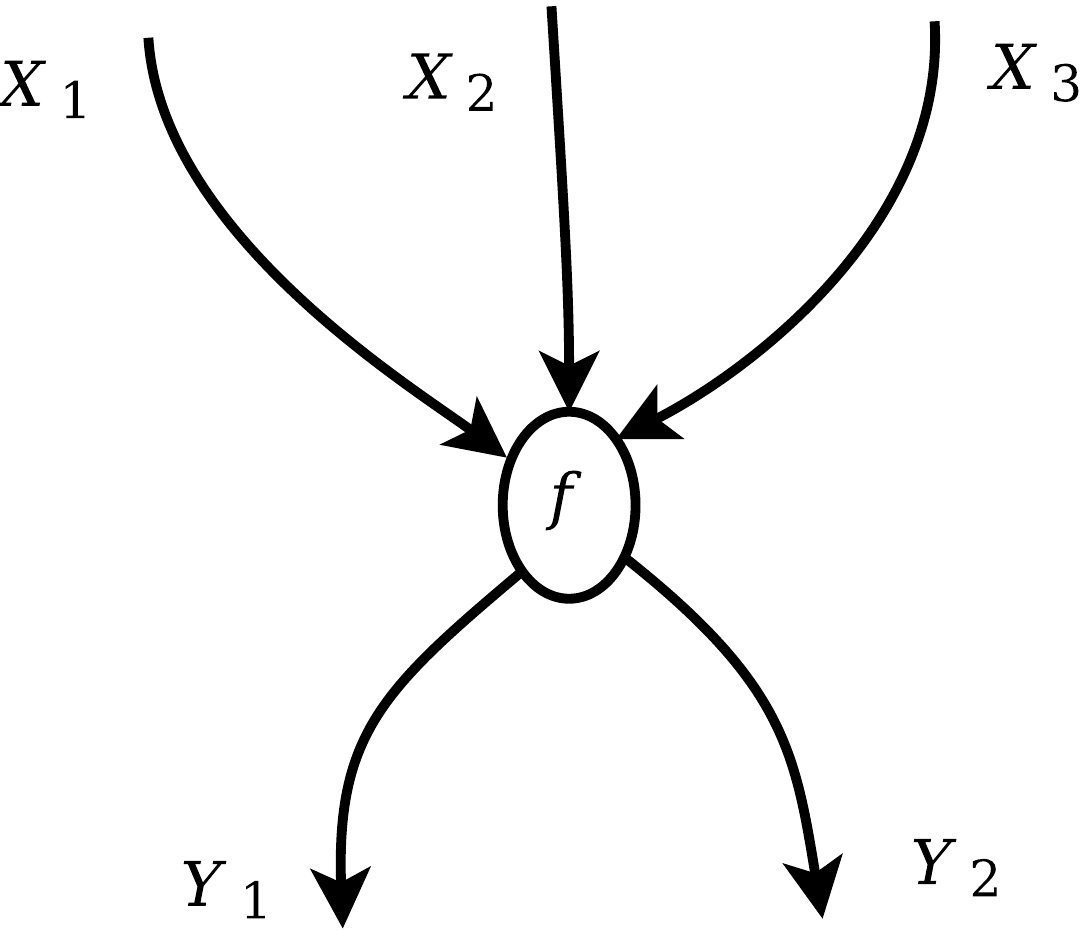}

\end{figure}

It has been widely noted that tensor algebras have the structure of
monoidal categories; perhaps the most insightful and carefully explained
such development is given by Baez and Stay\cite{Baez2009}. The diagram
of a tensor shown above is taken from that paper; it is a diagrammatic
representation of a morphism $f:X_{1}\otimes X_{2}\otimes X_{3}\to Y_{1}\otimes Y_{2}$.
There are several interesting operations one can do with tensors.
One of them is the contraction of indexes between two tensors. For
example, to multiply a matrix $M_{ik}$ by a vector $v_{k}$, one
sums over the index $k$ to obtain another vector: $w_{i}=\sum_{k}M_{ik}v_{k}$.
The matrix $M_{ik}$ should be understood as a 2-tensor, having two
connectors, while vectors are 1-tensors. The intent here is that $M_{ik}$
is to be literally taken as a seed, with $M$ the germ, and $i$ and
$k$ the connectors on the germ. The vector $v_{k}$ is another seed,
with germ $v$ and connector $k$. The inner product $\sum_{k}M_{ik}v_{k}$
is a connected section. The multiplication of vectors and matrices
is the act of connecting together connectors to form links: multiplication
is linking.

Tensors have additional properties and operations on them, the most
important of which, for analysis, is their linearity. For the purposes
here, the linearity is not important, whereas the ability to contract
indexes is. The contraction of indexes, that is, the joining together
of connectors to form links, gives tensor algebras the structure of
a monoidal category. This is a statement that seems simple, and yet
carries a lot of depth. As noted above, the beta-reduction of lambda
calculus also looks like the joining together of connectors. This
is not accidental; rather, it is the side effect of the fact that
the internal language of closed monoidal categories is simply typed
lambda calculus. The words ``simply typed'' are meant to convey
that there is only one type. For the above example morphism, that
would mean that $X_{1}$ and $X_{2}$ and so on all have the same
type: $X_{1}=X_{2}=X_{3}=Y_{1}=Y_{2}$. The end-points on the seed
are NOT labeled; equivalently, they all carry the same label. This
is in sharp contrast to the earlier example\\

\qquad{}%
\begin{minipage}[t]{0.8\columnwidth}%
\textsf{is: this- \& example+;}%
\end{minipage}\\
\\
where the two connectors are labeled, and have different types, which
sharply limit what they connect to. The \textsf{this-} connector has
the type ``\textsf{this-is}'', and can only attach to another connector
having the same type, namely, the \textsf{is+} connector on ``this''\\

\qquad{}%
\begin{minipage}[t]{0.8\columnwidth}%
\textsf{this: is+;}%
\end{minipage}\\
\\
It may seem strange to conflate the concept of tensors and monoidal
categories with linguistic analysis, yet this has an rich and old
history, briefly touched on in the next section. The core principle
driving this is that the Lambek calculus, underpinning the categorial
grammars used in linguistic analysis, can be embedded into a fragment
of non-commutative linear logic. The remaining step is to recall that
linear logic is the logic of tensor categories; the non-commutative
aspect is a statement that the left and right products must be handled
distinctly.

\subsection*{Similar concept: Lambek Calculus}

The foundations of categorial grammars date back to Lambek in 1961\cite{Lambek61,Marcus1967}
and the interpretation in terms of tensorial categories proliferates
explosively in modern times. One direct example can be found in works
by Kartsaklis\cite{Kart2013,Kart2014}, where one can find not only
a detailed development of the tensorial approach, together with its
type theory, but also explicit examples, such as the tensor 
\[
\overrightarrow{men}\otimes\overline{built}\otimes\overrightarrow{houses}
\]
together with explicit instructions on how to contract this with a
different tensor 
\[
\mathcal{F}\left(\alpha_{\mbox{subj verb obj}}\right)=\epsilon_{W}\otimes1_{W}\otimes\epsilon_{W}
\]
to obtain the ``quantization'' of the sentence ``men built houses''.
This notation will not be explained here; the reader should consult
\cite{Kart2013} directly for details. The point to be made is that
this kind of tensorial analysis can be, and is done, and often invokes
words like ``quantum'' and ``entanglement'' to emphasize the connection
to linear logic and to linear type theory.

Unfortunately, it is usually not clearly stated that it is only a
fragment of linear logic and linear type theory that applies. In linguistics,
it is not the linearity that is important, but rather the conception
of frames (in the sense of Kripke frames in proof theory). Frames
have the important property of presenting choices or alternatives:
one can have either this, or one can have that. The property of having
alternatives is described by intuitionistic logic, where the axiom
of double-negation is discarded. This either-or choice appears as
the concept of a ``multiverse'' in quantum mechanics, and far more
mundanely as alternative parses in linguistics.

Another worthwhile example of tensor algebra can be found in equation
13 of \cite{Kart2014}, reproduced below: 
\[
\overline{verb}=\sum_{i}\left(\overrightarrow{subject}_{i}\otimes\overrightarrow{object}_{i}\right)
\]
where $\overrightarrow{subject}_{i}$ and $\overrightarrow{object}_{i}$
are meant to be the $i$th occurrence of a subject/object pair in
an observed corpus. If the corpus consisted of two sentences, ``a
banana is a kind of fruit'' and ``this apple is green'', then one
would write
\[
\overline{verb}=\left(\overrightarrow{banana}\otimes\overrightarrow{fruit}\right)+\left(\overrightarrow{apple}\otimes\overrightarrow{green}\right)
\]
where the verb, in this case, is ``is''. The control over the word
order, that is, the left-right placement of the dependencies, is controlled
by means of the pregroup grammar. The pregroup grammar and its compositionality
properties follow directly from the properties of the left-division,
right-division and multiplication in the Lambek calculus. A quick
modern mathematical review of the axioms of the Lambek calculus can
be found in Pentus\cite{Pentus98}, which also provides a proof of
equivalence to context-free grammars.

\subsection*{Similar concept: history and Bayesian inference}

Some first-principles applications of Bayesian models to natural language
explicitly make use of a sequential order, called the ``history''
of a document.\cite{Rosen1996} That is, the probability of observing
the the $n$-th word of a sequence is taken to be $P(w_{n}|h)$ where
$h=w_{n-1},w_{n-2},\cdots,w_{1}$ is termed the history. This conception
of probability is sharply influenced by the theory of Markov processes
and finite-state machines, dating back to the dawn of information
theory.\cite{Ash1965} In a finite-state process model, the future
state is predicated only on the current state, and thus the Markov
assumption holds. In deciphering such a process, one might not know
how the current state is correlated to the output symbol, thus leading
to the concept of a Hidden Markov Model (HMM). The concept of ``history''
is well-suited for such analysis. Several issues, however, make this
approach impractical for many common problems, including natural language.

\begin{figure}[h]
\caption{The history of a text as a sequence of words}
\includegraphics[width=0.4\columnwidth]{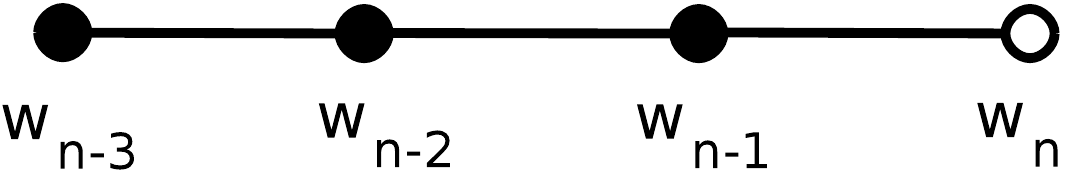}

\end{figure}

One issue, already noted, is the sequential nature of the process.
One can try to hand-wave away this issue: given a graph of vertices,
it is sufficient to write the vertexes in some order, any order will
do. This obscures the fact that $n$ vertexes have $n!$ ($n$-factorial)
possible interactions: a combinatorial explosion, when the actual
data graph may have a much much smaller number of interactions between
vertexes (aka ``edges''). By encoding the known interactions as
edges, a graphical approach avoids such a combinatorial explosion
from the outset.

To put it more bluntly: a sequential history model of genomic and
proteomic data is inappropriate. Although base pairs and amino acids
come in sequences, the interactions between different genes and proteins
are not in any way shape or form sequential. The interactions are
happening in parallel, in distinct, different physical locations in
a cell. These interactions can be depicted as a graph. Curiously,
that graph can resemble the one depicted below, although the depiction
is meant to show something different: it is meant to show a history.

\begin{figure}[h]
\caption{A Viterbi parse lattice of a Markov chain\label{fig:A-Viterbi-parse}}

\includegraphics[width=0.2\paperwidth]{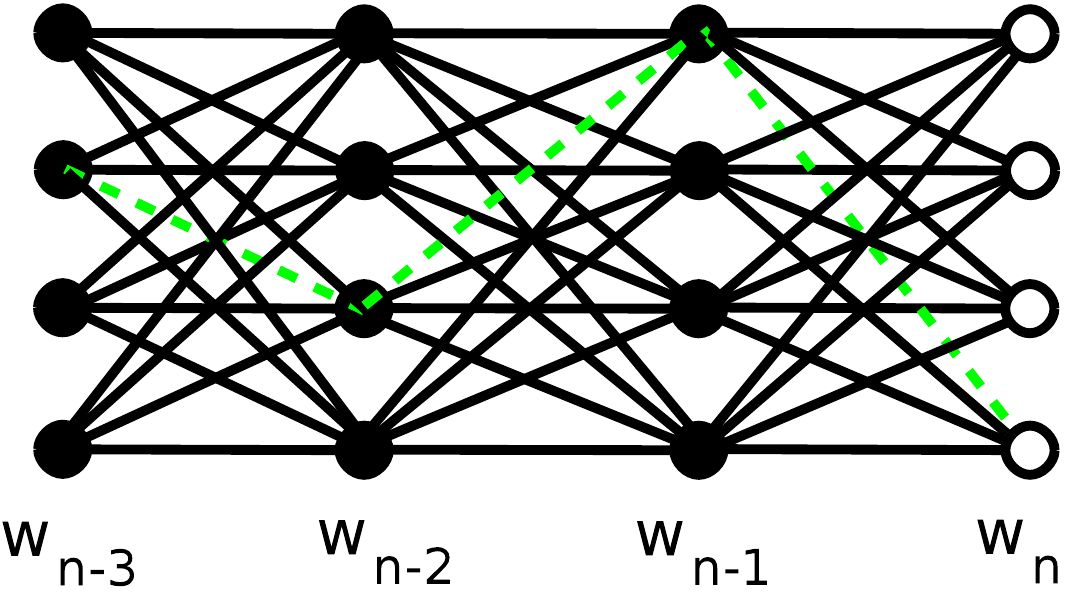}
\end{figure}

Figure \ref{fig:A-Viterbi-parse} depicts the lattice of a Viterbi
parse of a Markov chain. The dashed green line depicts a maximum-likelihood
path through the lattice, that is, the most likely history. Viterbi
decoding, using an ``error correcting code'', is a process by which
the validity of the dashed green path is checked, and failing paths
discarded. For natural language, the dashed red path must be a grammatically
correct sequence of words. For a radio receiver, the dashed red path
must be a sequence of bits that obey some error-correction polynomial;
if it doesn't, the next-most-likely path is selected.

Each black line represents a probability $p_{ij}$ of moving from
state $i$ to state $j$ at the next time-step. That is, $p_{ij}=P(w_{n}=j|w_{n-1}=i)$
is the likelihood of word $j$ given word $i$ in the immediate past.
The probabilities are arranged such that $\sum_{i}p_{ij}=1$. This
is called a Markov model, because only the most recent state transitions
are depicted: there are no edges connecting the nodes more than one
time-step apart; there are no edges connecting $w_{n}$ to $w_{n-2},$
etc. Put differently, $P(w_{n}|h)=P(w_{n}|w_{n-1})$. That is, this
depicts the use of 2-grams to predict the current state.

Non-Markov models would have edges connecting nodes further in the
past. A $n$-gram approach to language digs $n$ steps into the past.
If there are $k$ states, and $n$ steps into the past, then $k^{n}$
edges are required: that is, a rank-$n$ tensor. Here, $k=4$ and
$n=2$ is depicted; in natural language $k$ is the number of words
(say, $k=10^{4}$ for a common subset of the English language), while
$n$ is the length of a longer sentence, say $n=12$. In this case,
the history tensor $P(w_{n}|h)$ has $k^{n}=10^{48}=2^{160}$ edges.
But of course, this is computationally absurd. It is also theoretically
absurd: almost all of those edges have zero probability. Almost none
of the edges are needed; the actual tensor is very very sparse. 

The red path in the figure below indicates a very unlikely word-sequence:
``example this an this''. There are $4\times16=64$ paths through
it. Of these, only 3 are plausible: the green edges, and the sequences
``this example is an'' and ``an example is this''. The others
can't be observed.

\begin{figure}[h]
\caption{Likely and unlikely word sequences}

\includegraphics[width=0.35\columnwidth]{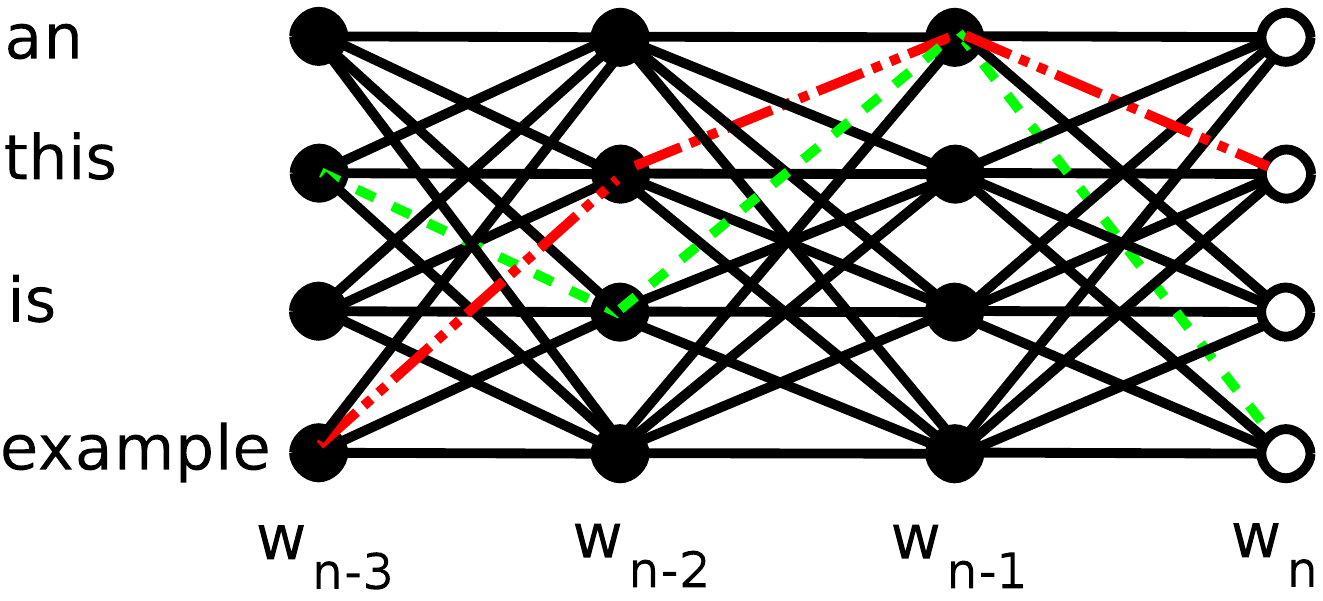}

\end{figure}

The sparsity is easily exposed with dependency parsing. So, for example,
if $w_{n-3}=this$ and $w_{n-2}=is$ and $w_{n-1}=an$, a dependency
parse will tell you that $w_{n}$ must be a singular noun starting
with a vowel, or an adjective starting with a vowel. It also tells
you that, for this particular history, this noun can depend only on
$w_{n-2}$ and on $w_{n-1}$ but not on $w_{n-3}$. A collection of
dependency parses obtained from a corpus identifies which edges matter,
and which edges do not. 

Dependency parses do even more: they unveil possible paths, and not
just pair-wise edges. They provide a more holistic view of what might
be going on in natural language. That is, the notation 

\[
\overline{is}=\left(\overrightarrow{banana}\otimes\overrightarrow{fruit}\right)+\left(\overrightarrow{apple}\otimes\overrightarrow{green}\right)
\]
and\\

\qquad{}%
\begin{minipage}[t]{0.8\columnwidth}%
\textsf{is: (banana- \& fruit+) or (apple- \& green+);}%
\end{minipage}\\
\\
and 
\[
P\left(w_{n}=fruit|w_{n-1}=is,w_{n-2}=banana\right)+P\left(w_{n}=green|w_{n-1}=is,w_{n-2}=apple\right)
\]
all represent the same knowledge, the dependency notation appears
to be less awkward than thinking about history as some Bayesian probability.
The dependency notation focuses attention on a different part of the
problem. 

Another popular way to at least partly deal with the sparsity of the
history tensor $P(w_{n}|h)$ is to use skip-grams. The idea recognizes
that many of the edges of an $n$-gram will be zero, and so these
edges can be skipped. This is not a bad approach, except that it is
``simply typed'': it does not leverage the possibility that different
words might have different types (verb, noun, ...) and that this typing
information delivers further constraints on the structure of the skip-gram.
That is, the notion of subj-verb-obj not only tells you that your
skip-gram is effectively a 3-gram, but also that the first and third
words belong to a class called ``noun'', and the middle is a transitive
verb. This sharply prunes the number of possibilities \emph{before}
the learning algorithm is launched, instead of during or after. The
fact that such pruning is even possible is obscured by the notation
and language of $n$-grams and the history $P(w_{n}|h)$. 

A different stumbling block of the ``history'' approach is that
it ignores ``the future'': the fact that the words that might be
said next have already influenced the choice of the words already
spoken. This can be hand-waved away by stating that the history is
creating a model of (hidden) mental states, and that this model already
incorporates those, and thus is anticipating future speech actions.
Although this might be philosophically acceptable to some degree,
it again forces complexity onto the problem, when the complexity is
not needed. If you've already got the document, look at all of it;
go all the way to the end of the sentence. Don't arbitrarily divide
it into past and future, and discard the future.

To summarize: dependency structures appear naturally; flattening them
into sequences places one at a notional, computational and conceptual
disadvantage, even if the flattening is conceptually isomorphic to
the original problem. The tensor $P(w_{n}|h)$ may indeed encode all
possible knowledge about the text in a rigorously Bayesian fashion;
but its unwieldy.

\section*{Quotienting}

The intended interpretation for the graphs discussed in this document
is that they represent or are the result of capturing a large amount
of collected raw data. From this data, one wants to extract commonalities
and recurring patterns. 

The core assumption being made in this section is that, when two local
neighborhoods of a graph are similar or identical, then this reflects
some important similarity in the raw data. That is, similarity of
subgraphs is the be-all and end-all of extracting knowledge from the
larger graph, and that the primary goal is to search for, mine, such
similar subgraphs.

Exactly what it means to be ``similar'' is not defined here; this
is up to the user. Similarity could mean subgraph isomorphism, or
subgraph homomorphism, or something else: some sort of ``close-enough''
similarity property involving the shape of the graph, the connections
made, the colors, directions, labels and weights on the vertexes or
edges. The precise details do not matter. However, it is assumed that
the user can provide some algorithm for finding such similarities,
and that the similarities can be understood as a kind-of ``equivalence
relation''.

\subsection*{Example of similarity}

To motivate this, consider the following scenario. One has a large
graph, some dense mesh, and one decides, via some external decision
process, that two vertexes are similar. One particularly good reason
to think that they are similar is that they share a lot of nearest
neighbors. In a social graph, one might say they have a lot of friends
in common. In genomic or proteomic data, they may interact with the
same kinds of genes/proteins. In natural language, they might be words
that are synonyms, and thus get used the same way across many different
sentences; specifically, the syntactic dependency parse links these
words to the same set of heads and dependents. At any rate, one has
a large graph, and some sort of equivalence operation that can decide
if two vertexes are the ``same'', or are ``similar enough''. Whenever
one has an equivalence relation, one can apply it to obtain a quotient,
of grouping together into an identity all things that are the same.

To make this even more concrete, consider this example from linguistics.
Suppose, given a corpus, one has observed three sentences: ``Mary
walked home'', ``Mary ran home'' and ``Mary drove home''. A dependency
parse provides three seeds: \\

\qquad{}%
\begin{minipage}[t]{0.8\columnwidth}%
\textsf{walked: Mary- \& home+; }

\textsf{ran: Mary- \& home+;}

\textsf{drove: Mary- \& home+;}%
\end{minipage}\\
\\
which seem to be begging for an equivalence relation that will reduce
these to \\

\qquad{}%
\begin{minipage}[t]{0.8\columnwidth}%
\textsf{walked ran drove: Mary- \& home+; }%
\end{minipage}\\
\\
Using a tensorial notation, once starts with
\[
\overrightarrow{Mary}\otimes\overline{walked}\otimes\overrightarrow{home}+\overrightarrow{Mary}\otimes\overline{ran}\otimes\overrightarrow{home}+\overrightarrow{Mary}\otimes\overline{drove}\otimes\overrightarrow{home}
\]
and applies the equivalence relation to obtain

\[
\overrightarrow{Mary}\otimes\left(\overline{walked}+\overline{ran}+\overline{drove}\right)\otimes\overrightarrow{home}
\]
The structure here strongly resembles the application of the distributive
law of multiplication over addition. This distributivity property
is one of the appeals of the tensor notation. One can obtain a similar
sense of distributivity by using the operator ``\textsf{or}'' to
separate the Link Grammar style stanzas, and note that the change
also appears to be an application of the distributive law of conjunction
over disjunction. 

\begin{figure}[h]
\caption{Creating a quotient graph\label{fig:Creating-a-quotient}}
\includegraphics[width=0.55\columnwidth]{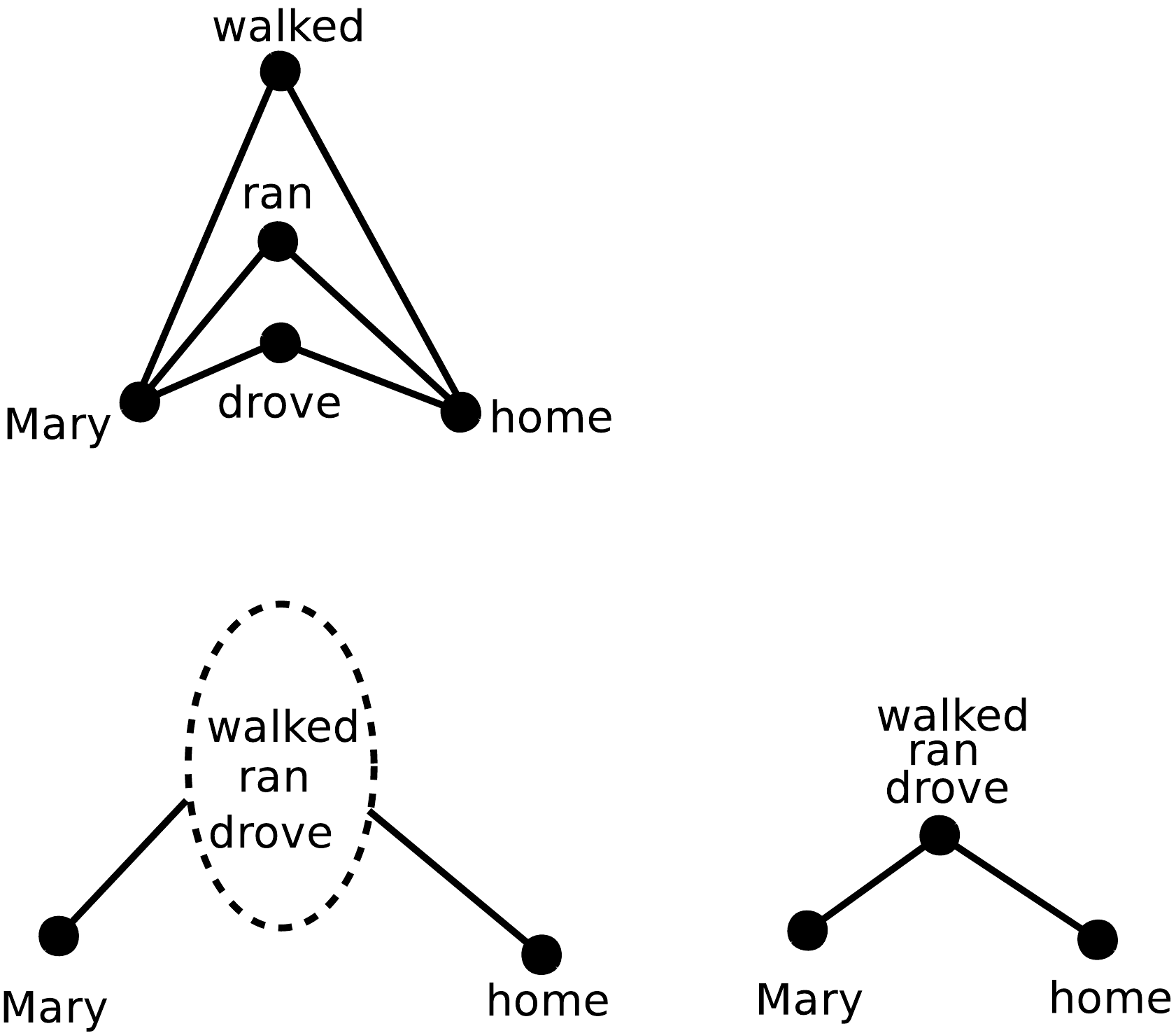}
\end{figure}

This is illustrated pictorially, in figure \ref{fig:Creating-a-quotient}.

It need not be the case that an equivalence relation is staring us
in the face, yet here, it is. The vertexes ``walked'', ``ran''
and ``drove'' can be considered similar, precisely because they
have the same neighbors. The upper graph can be simplified by computing
a quotient, shown in the lower part: the quotient merges these three
similar vertexes into one. The result is not only a simpler graph,
but also some vague sense that ``walked'', ``ran'' and ``drove''
are synonymous in some way.

\subsection*{Quotienting}

If one has an equivalence relation that can be applied to a graph,
then the obvious urge is to attempt to perform quotienting on the
graph. That is, to create a new graph, where the ``equal'' parts
are merged into one. 

The first issue to be cleared out of the way is the use of the word
``\href{https://en.wikipedia.org/wiki/Quotient}{quotienting}'',
which seems awkward, since the example above seemed to involve some
sort of factoring, or the application of a distributive law of some
sort. The terminology comes from modulo arithmetic, and is in wide
use in all branches of mathematics. A simple example is the idea of
dividing by three: given the set of integers $\mathbb{Z}$, one partitions
it into three sets: the set $\left\{ 0,3,6,9,\cdots\right\} $, the
set $\left\{ 1,4,7,\cdots\right\} $ and the set $\left\{ 2,5,8,\cdots\right\} $.
These three sets are termed the cosets of 0, 1 and 2, and all elements
in each set are considered to be equal, in the sense that, for any
$m$ and $n$ in any one of these sets, it is always true that $m=n\mod3$:
they are equal, modulo 3. In this way, one obtains the quotient set
$\mathbb{Z}_{3}=\mathbb{Z}/3\mathbb{Z}=\mathbb{Z}/\mod3=\left\{ 0,1,2\right\} $.
Modulo arithmetic resembles division, ergo the term ``quotient''. 

Given a set $S$ and an equivalence relation $\sim$, it is common
to write the quotient set as $Q=S/\sim$. In the above, $S$ was $\mathbb{Z}$
and $\sim$ was $\mod3$. In general, one looks for, and works with
equivalence relations that preserve desirable algebraic properties
of the set, while removing undesirable or pointless distinctions.
In the modulo arithmetic example, addition is preserved: it is well
defined, and works as expected. In the linguistic example, the subj-verb-obj
structure of the sentence is preserved; the quotienting removes the
``pointless'' distinction between different verbs.

Quotienting is often described in terms of homomorphisms, functions
$\pi:S\to Q$ that preserve the algebraic operations on $S$. For
example, if $m:S\times S\times S\to S$ is a three-argument endomorphism
on $S$, one expects that $\pi$ preserves it: that $\pi\left(m\left(a,b,c\right)\right)=m\left(\pi\left(a\right),\pi\left(b\right),\pi\left(c\right)\right)$.
For the previous example, if $m$ was used to provide or identify
a subj-verb-obj relationship, then, after quotienting, one expects
that $m$ can still identify the verb-slot correctly. 

\subsection*{Graph quotients}

In graph theory, the notion of quotienting is often referred to as
working ``relative to a subgraph''. Given a graph $G$ and a subgraph
$A\subset G$, one ``draws a dotted line'' or places a balloon around
the vertexes and edges in $A$, but preserves all of the edges coming
out of $A$ and going into $G$. The internal structure of $A$ is
then ignored. The equivalence relation makes all elements of $A$
equivalent, so that $A$ behaves as if it were a single vertex, with
assorted edges attached to it, running from $A$ to the rest of $G$.

\subsection*{Stalks}

Given the above notion of a graph quotient, it can be brought over
to the language of seeds and sections, established earlier. Let $G$
be a graph, and let $v_{a}$ and $v_{b}$ be two vertexes in the graph,
with corresponding seeds $s_{a}$ and $s_{b}$ extracted from the
graph. That is, $s=\left(v,C_{v}\right)$ with $C_{v}$ being the
set of edges connecting $v$ to all of its nearest neighbors. Let
$\pi$ be a projection function, such that $\pi\left(v_{a}\right)=\pi\left(v_{b}\right)$.
That is, $\pi:V\to B$ is a map from the vertices $V$ of $G$ to
some other set $B$. 

It is not hard to see that $\pi$ is a morphism of graphs; it not
only maps vertexes, but it can be extended to map edges as well. The
target of $\pi$ is a graph quotient.
\begin{defn*}
Given a map $\pi:V\to B$, the \noun{stalk} above $b\in B$ is the
set $S$ of seeds such that for each $s=\left(v,C_{v}\right)\in S$,
one has that $\pi(v)=b$. $\diamond$
\end{defn*}
\begin{figure}[h]
\caption{A stalk and it's projection}
\includegraphics[width=0.6\columnwidth]{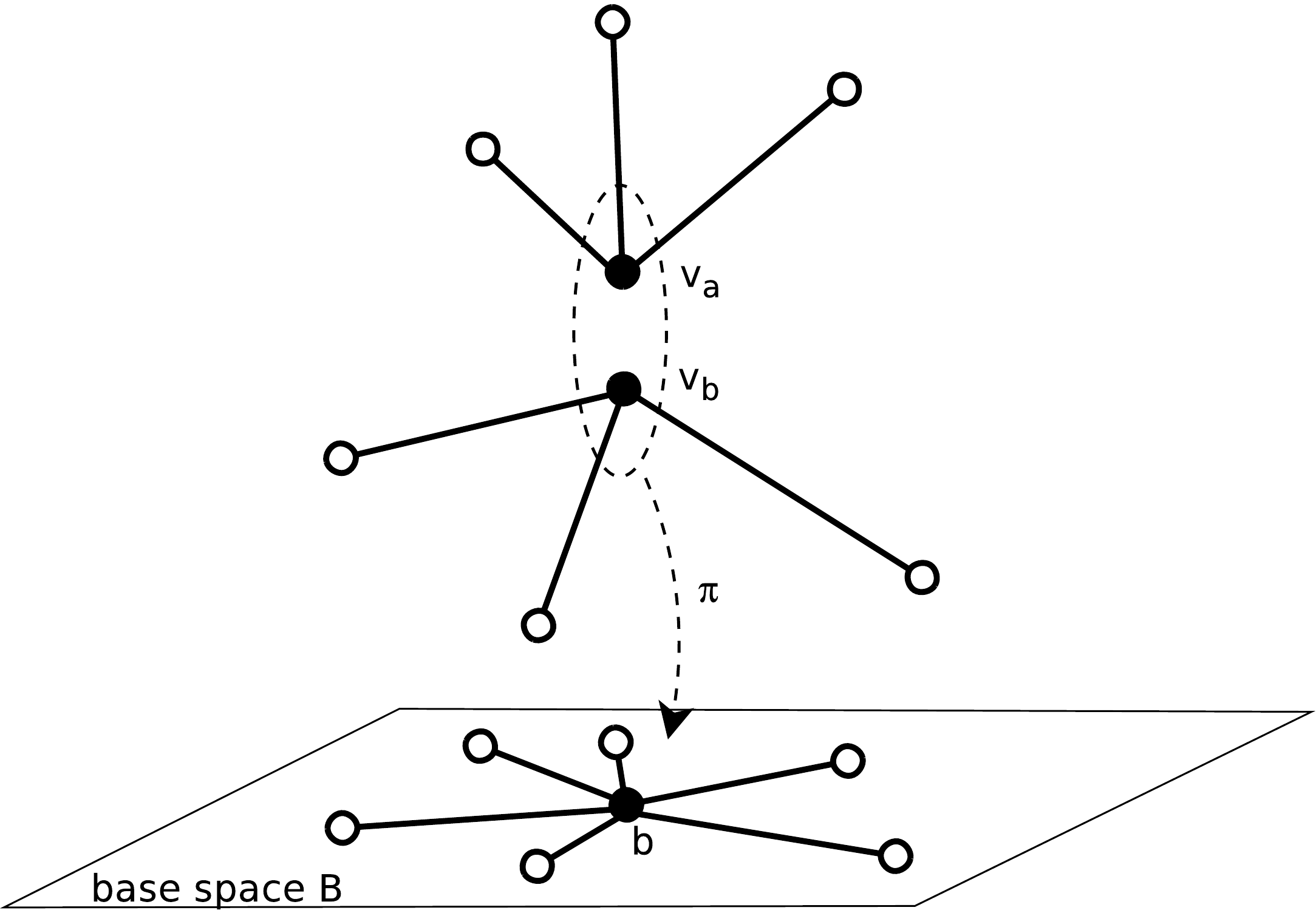}
\end{figure}

In general, this definition does not require that the map $\pi:V\to B$
be a total map; that is, it does not need to be defined on all of
$V$. Also, $V$ does not need to be the vertexes of some specific
graph; it is enough that $V$ is a set of germs of seeds. That is,
the seeds in the stalk can be generalized seeds, having typed connectors,
rather than connectors derived from edges. The vertexes in the stalk
can be visualized as being stacked one on top another, forming a tower
or a fiber, with the edges sticking out as spines. When the seeds
carry typed connectors, the stalk can be visualized as a tower of
jigsaw-puzzle pieces. 

\begin{figure}[h]
\caption{A corn stalk, a stack of puzzle pieces}
\includegraphics[width=0.25\textwidth]{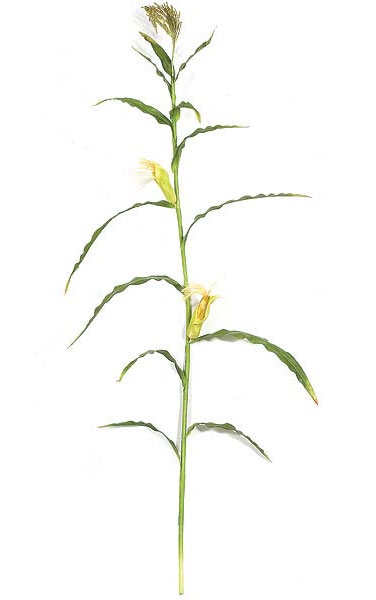}\includegraphics[width=0.25\columnwidth]{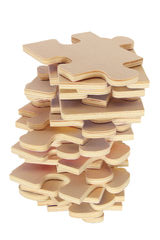}

\end{figure}

Note that the projection of a stalk is a seed. It's germ is $b$,
and if any connector appears in the stalk, then it also appears as
a connector on $b$ in the base. At least, this is the unassailable
conclusion if one starts with a graph, and assumes that $\pi$ is
a graph morphism. It will prove to be very useful to loosen this restriction,
that is, to allow $\pi$ to add or remove connectors. Thus, it is
useful to immediately broaden the definition of the stalk. 
\begin{defn*}
Given a map $\pi:E\to B$, where both $E$ and $B$ are collections
seeds, the \noun{stalk} above $b\in B$ is the set $S$ of seeds in
$E$ such that for each $s=\left(v,C_{v}\right)\in S$, one has that
$\pi(s)=b$. $\diamond$
\end{defn*}
In this revised definition, there is no hint of what $\pi$ did with
the connectors. In particular, there is no way to ask about some specific
connector on some seed $s$, and what happened to it after $\pi$
mapped $s$ to $b$. This definition is perhaps too general; in the
most common case, it is useful to project the connectors as well as
the germs. It is also very useful to be able to say that a particular
connector on $s$ can be mapped to a particular connector on $b$.
Yet it is also useful to sometimes discard some connectors because
they are infrequently used, to perform pruning, as it were. These
use-cases will be returned to later. There is no particular reason
to allow pruning during projection; it can always be done before,
or after.

Thus, perhaps the most agreeable definition for a stalk is this.
\begin{defn*}
Given a map $\pi:E\to B$, where both $E$ and $B$ are collections
seeds, the \noun{stalk} above $b\in B$ is the set $S$ of seeds in
$E$ such that for each $s=\left(v,C_{v}\right)\in S$, one has that
$\pi(s)=b$. The map $\pi$ can be decomposed into a pair $\pi=\left(\pi_{g},\pi_{c}\right)$
such that, for every $\gamma\in C_{v}$ one has that $\pi\left(v,\gamma\right)=\left(\pi_{g}\left(v\right),\pi_{c}\left(\gamma\right)\right)$
such that $\pi_{c}\left(\gamma\right)\in C_{b}$. That is, $\pi_{g}$
maps the germs of $E$ to the germs of $B$ and $\pi_{c}$ maps the
connectors in $E$ to specific connectors in \textbf{$B$.} $\diamond$
\end{defn*}
The next figure illustrates both the projection of germs, and of connectors.
It tries to capture the notion that the projection is entire and consistently
defined.

\begin{figure}[h]

\caption{Germs and connectors project consistently}

\includegraphics[width=0.6\columnwidth]{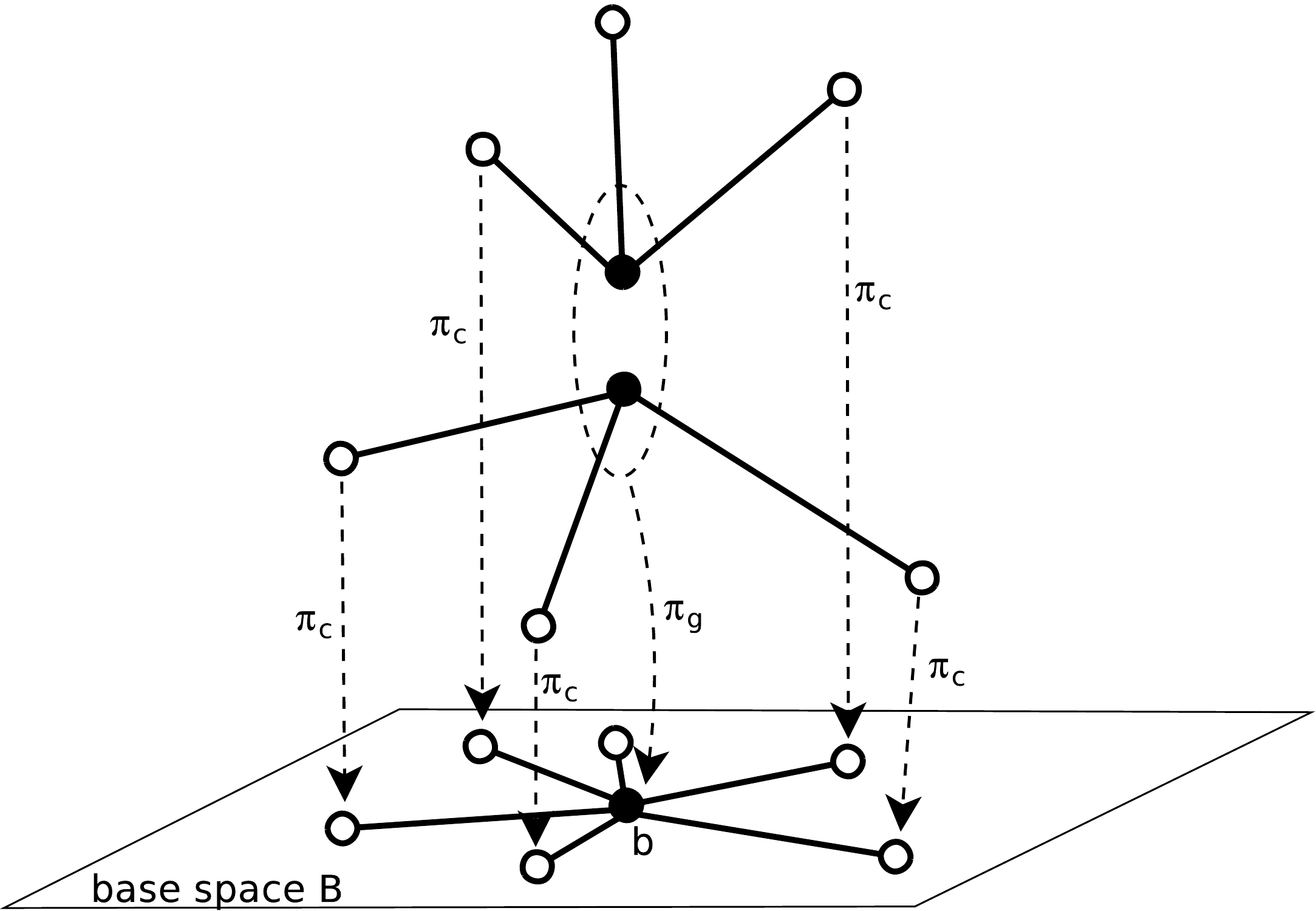}

\end{figure}

The definition of a link needs to be generalized, and made consistent
with this final definition of a stalk. 
\begin{defn*}
Two stalks $S_{1}$ and $S_{2}$ are \noun{connected} if there exists
a link between some seed $s_{1}\in S_{1}$ and some seed $s_{2}\in S_{2}$.
The stalks are \noun{consistently linked} if the projections of the
stalks are also linked in a fashion consistent with the projection.
That is, if $\left(v_{1},t_{a}\right)$ is the connector on $s_{1}$
that is connected to the connector $\left(v_{2},t_{b}\right)$ on
$s_{2}$, \emph{viz.} $v_{2}\in t_{a}$ and $v_{1}\in t_{b}$, then
$\left(\pi_{g}\left(v_{1}\right),\pi_{c}\left(t_{a}\right)\right)$
is connected to $\left(\pi_{g}\left(v_{1}\right),\pi_{c}\left(t_{a}\right)\right)$
. That is, $\pi_{g}\left(v_{2}\right)\in\pi_{c}\left(t_{a}\right)$
and $\pi_{g}\left(v_{1}\right)\in\pi_{c}\left(t_{b}\right)$ .$\diamond$
\end{defn*}
Recall that the original definition of a connector was such that it
could be used once and only once. This can become an issue, if it
is strictly enforced on the base space. It will become convenient
to remove this restriction on the base space, and replace it by a
use-count. That is, if two different links between stalks project
down to the same link in the base space, then the link in the base-space
should be counted ``with multiplicity''. This induces the notion
that maybe the base space can be used for statistics-gathering, and
that is exactly the intent. 

\section*{Sheaves}

The stalk is meant to provide a framework with which to solve the
computational intractability problems associated with Bayesian networks,
by explicitly exposing the grammatical structure within them in such
a fashion that they can be explicitly manipulated. The intent is to
accomplish the hope expressed in the diagram below. To actually arrive
at a workable solution requires additional clarifications, examples,
and definitions. This hopeful figure \emph{must not be taken literally}:
one certainly does \emph{not} want the base space to be some Markov
network! That would be a disaster. Rather, the hope is to accumulate
a large number of graph fragments in such a way that the fragments
are apparent, but that the statistics of their collective behavior
is also accessible. The hope is that this can be done without overflowing
available CPU and RAM, while carefully maintaining fidelity to the
graph fragments. This is an example from linguistics, but one might
hope to do the same with activation pathways in cell biochemistry.
The citric acid cycle should be amenable to such a treatment, as well.

\begin{figure}[h]

\caption{The problem, and it's intended solution }

\includegraphics[width=0.45\columnwidth]{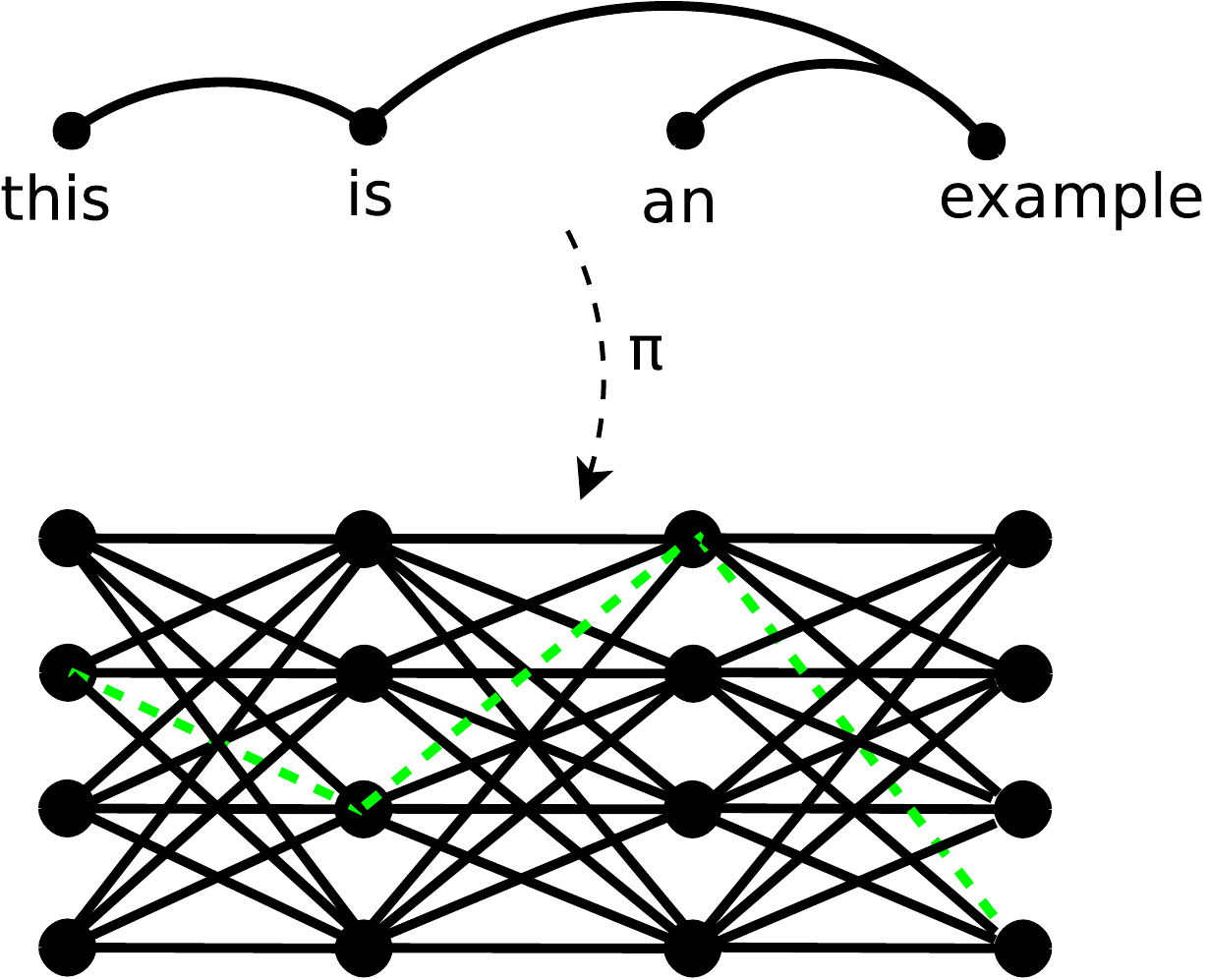}

\end{figure}

From the previous development, it should be clear that stalks capture
the local structure of graphs, and that the projection, carefully
done, can preserve the essence of that local structure. Enough mechanism
has been developed to allow the definition of a section to be understood
in a way that is in keeping with the usual notion of a section as
commonly defined in covering spaces and fiber bundles. A preliminary,
provisional definition of a sheaf can now be given.
\begin{defn*}
A sheaf is a collection of connected sections, together with a projection
function $\pi$ that can be taken to be an equivalence relation. That
is, $\pi$ maps sections to a base space $B$, such that, for each
pair of vertexes $v,w$ occurring in different sections, one has $\pi(v)=\pi(w)$
if and only if $v,w$ are in germs in the same stalk. $\diamond$
\end{defn*}
This provisional definition can be tightened. The formal definition
of a sheaf also requires that it obey a set of axioms, called the
gluing axioms. Before giving these, it is useful to look at an example.

\subsection*{Example: collocations}

A canonical first step in corpus linguistics is to align text around
a shared word or phrase:\medskip{}

\qquad{}%
\noindent\begin{minipage}[t]{1\columnwidth}%
\begin{tabular}{rl}
 & \texttt{fly like a butterfly}\tabularnewline
\texttt{airplanes that} & \texttt{fly}\tabularnewline
 & \texttt{fly fishing}\tabularnewline
 & \texttt{fly away home}\tabularnewline
 & \texttt{fly ash in concrete}\tabularnewline
\texttt{when sparks} & \texttt{fly}\tabularnewline
\texttt{let's} & \texttt{fly a kite}\tabularnewline
\texttt{learn to} & \texttt{fly helicopters}\tabularnewline
\end{tabular}%
\end{minipage} 

\medskip{}

Each word is meant to be a vertex; edges are assumed to connect the
vertexes together in some way. In standard corpus linguistics, the
edges are always taken to join together neighboring words, in sequential
fashion. Note that each phrase in the collocation obeys the formal
definition of a section, given above. It does so trivially: its just
a linear sequence of vertexes connected with edges. If the collocated
phrases are chopped up so that they form a word-sequence that is exactly
$n$ words long, then one calls that sequence an $n$-gram. 

The projection function $\pi$ is now also equally plain: it simply
maps all of the distinct occurrences of the word ``fly'' down to
a single, generic word ``fly''. The stalk is just the vertical arrangement
of the word ``fly'', one above another. Each phrase or section can
be visualized as a botanical branch or botanical leaf branching off
the central stalk.The projection of all of the stalks obtained from
collocation is shown below, in figure \ref{fig:N-gram-base}. Identical
words are projected down to a common base point. Links between words
are projected down to links in the base space. For ordinary $n$-grams,
the links are merely the direct sequential linking of neighboring
words. The figure depicts the base-space of the sheaf obtained from
$n$-grams.

\begin{figure}[h]
\caption{N-gram corpus text alignment\label{fig:N-gram-base}}

\includegraphics[width=0.5\columnwidth]{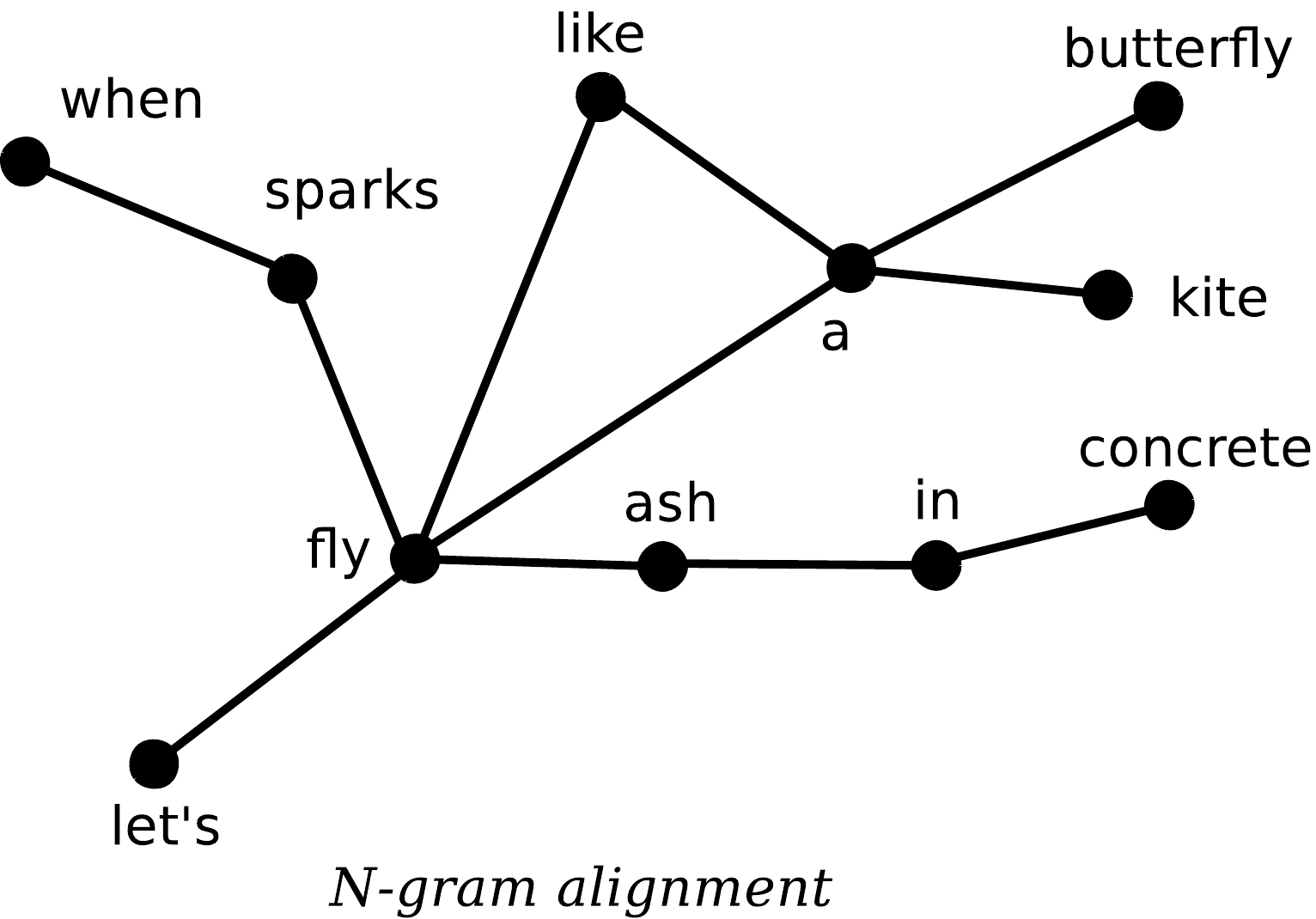}
\end{figure}

The sections do not have to be linear sequences; the phrases can be
parsed with a dependency parser of one style or another, in which
case the words are joined with edges that denote dependencies. The
edges might be directed, and they might be labeled. Parsing with a
head-phrase parser introduces additional vertexes, typically called
NP, VP, S and so on. The next figure (figure \ref{fig:Dependency-base})
shows the projection that results from alignment on an (unlabeled,
undirected) dependency parse of the text. As before, each stalk is
projected down to a single word, and the links are projected down
as well. The most noticeable difference between this base space and
the N-gram base space is that the determiner ``a'' does not link
to ``fly'' even though it stands next to it; instead, the determiner
links to the noun it determines. This figure also shows ``ash''
as modifying ``fly'', which, as a dependency, is not exactly correct
but does serve to illustrate how the N-gram and the dependency alignments
differ. If the dependency parse produced directed edges with labels,
it would be prudent to project those labels as well.

\begin{figure}[h]
\caption{Dependency parse corpus text alignment\label{fig:Dependency-base}}

\includegraphics[width=0.75\columnwidth]{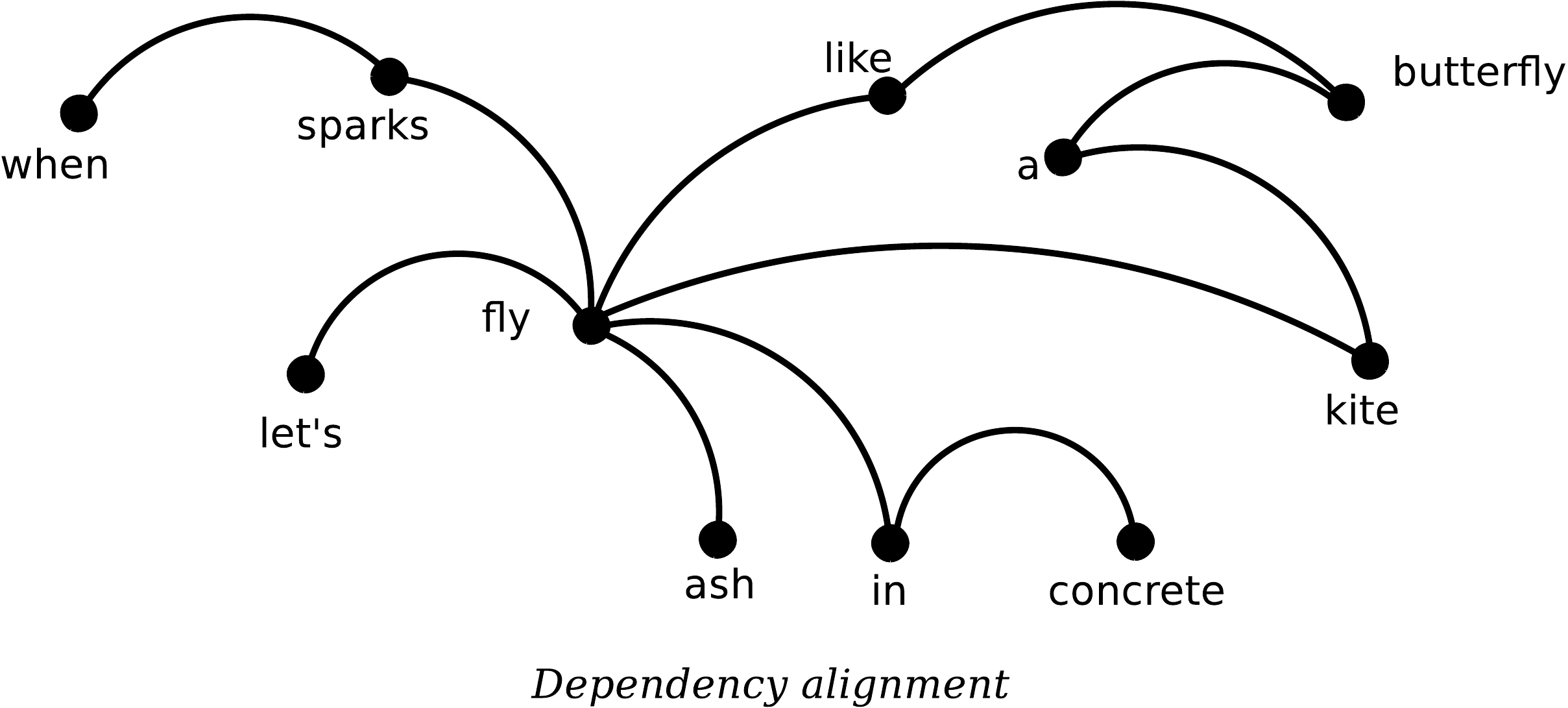}
\end{figure}

Both of the figures \ref{fig:N-gram-base} and \ref{fig:Dependency-base}
depict a quotient graph that results from a corpus alignment, where
all uses of a word have been collapsed (projected down) to a single
node, and all links connecting the words are likewise projected. The
resulting graph can be understood to depict all possible connections
in a natural language. In some sense, it captures important structural
information in natural language.

Be careful, though: these base spaces are just the projections of
the sheaf; they are not the sheaf itself. Its as if a flashlight were
held above the stalks: the base space is the shadow that is cast.
The sheaf is the full structure, the base space is just the shadow. 

\subsection*{Are projections useful?}

Yes. A collapsed graph like those above might appear strange; why
would one want to do that, if one has individual sentence data? 

By collapsing in this way, one obtains a natural place to store \href{https://en.wikipedia.org/wiki/Marginal_distribution}{marginal distributions}.
For example, when accumulating statistics for large collections of
sentences, the projected vertex becomes an ideal place to store the
frequency count of that word; the projected edge becomes an excellent
place to store the joint probability or the mutual information for
a pair of words. The projected graph - the quotient graph, is manageable
in size. For example, in a corpus consisting of ten million sentences,
one might see 130K distinct, unique words (130K vertexes) and perhaps
5 million distinct word-pairs (5M edges). Such a graph is manageable,
and can fit into the RAM of a contemporary computer. 

By contrast, storing the individual parses for 10 million sentences
is more challenging. Assuming 15 words per sentence, this requires
storing 150M vertexes, and approximately 20 links per sentence for
200M edges. This graph is two orders of magnitude larger than the
quotient graph. One could, of course, apply various programming and
coding tricks to squeeze and compress the data, but this misses the
point: It makes sense to project sections down to the base space as
soon as possible. The original sections can be envisioned to still
be there, virtually, in principle, but the actual storage can be avoided.

Every graph can be represented as an adjacency matrix. In this example,
it would be a sparse matrix, with 5 million non-zero entries out of
130K$\times$130K total. The sparsity is considerable: $\log_{2}\left(130\times130/5\right)=11.7$.
Less than one in a thousand of all possible edges are actually observed.

The marginals stored with the graph can be accessed as marginals on
the adjacency matrix. That is, they are marginals in the ordinary
sense of values written in the margin of the matrix. Standard linear-algebra
and data-analysis tools, such as the R programming language, can access
the matrix and the marginals.

\subsection*{Visualizing Sheaves}

One way of visualizing the sheaf is as a stack of sheets of paper,
with one sentence written on each sheet. The papers are stacked in
such a way that words that are the same are always arranged vertically
one above another. This stacking is where the term ``sheaf'' comes
from. Each single sheet of paper is a section. Each collocation is
a stalk. 

\begin{figure}
\caption{A Sheaf of Stalks; a Sheaf of Paper}

\includegraphics[width=0.35\columnwidth]{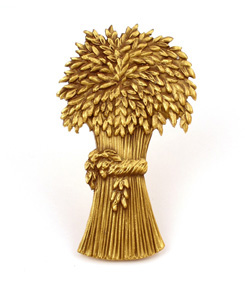}\includegraphics[width=0.51\columnwidth]{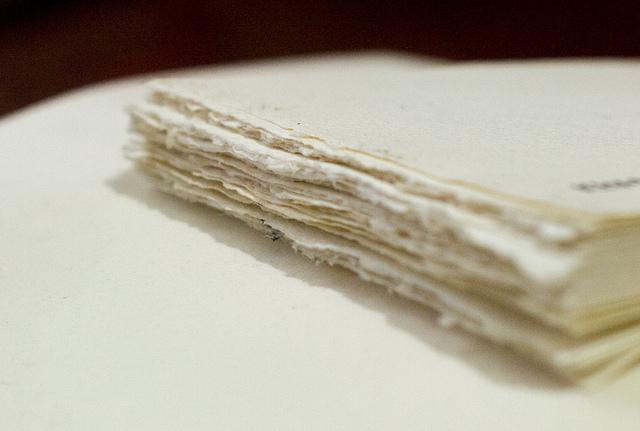}
\end{figure}

A different example can be taken from biochemistry. There, one might
want to write down specific pathways or interaction networks on the
individual sheets of paper, treating them as sections. If one specific
gene is up-regulated, one can then try to view everything else that
changed as belonging to the same section, as if it were an activation
mode within the global network graph of all possible interactions.
Thus, for example, the Krebs cycle can be taken to be a single section
through the network: it shows exactly which coenzymes are active in
aerobic metabolism. The same substrates, products and enzymes may
also participate in other pathways; those other pathways should be
considered as other sections through the sheaf. Each substrate, enzyme
or product is itself a stalk. Each reaction type is a seed.

The sheaf, it's decomposition into sections, and it's projection down
to a single unified base network, provides a holistic view of a network
of interactions. For linguistic data, activations or modes of the
network correspond to grammatically valid sentences. For biological
data, an activated biological pathway is a section. The base space
provides a general map of biochemical interactions; it does not capture
individual activations. The individual sections in the sheaf do capture
that activation.

\subsection*{Feature Vectors}

It is important to understand that, in many ways, stalks can be treated
as vectors, and, specifically as the ``feature vectors'' of data-mining.
This is best illustrated with an example.

Consider the corpus ``the dog chased the cat'', ``the cat chased
the mouse'', ``the dog chased the squirrel'', ``the dog killed
the chicken'', ``the cat killed the mouse'', ``the cat chased
the cockroach''. There are multiple stalks, here, but the ones of
interest are the one for the dog:

\medskip{}

\qquad{}%
\noindent\begin{minipage}[t]{1\columnwidth}%
\begin{tabular}{rl}
\texttt{the} & \texttt{dog chased the cat}\tabularnewline
\texttt{the} & \texttt{dog chased the squirrel}\tabularnewline
\texttt{the} & \texttt{dog killed the chicken}\tabularnewline
\end{tabular}%
\end{minipage} 

\medskip{}
and the stalk for the cat:

\medskip{}

\qquad{}%
\noindent\begin{minipage}[t]{1\columnwidth}%
\begin{tabular}{rl}
\texttt{the dog chased the} & \texttt{cat}\tabularnewline
\texttt{the} & \texttt{cat chased the mouse}\tabularnewline
\texttt{the} & \texttt{cat killed the mouse}\tabularnewline
\texttt{the} & \texttt{cat chased the cockroach}\tabularnewline
\end{tabular}%
\end{minipage} 

\medskip{}

One old approach to data mining is to trim these down to 3-grams,
and then compare them as feature vectors. These 3-gram feature vector
for the dog is:

\medskip{}

\qquad{}%
\noindent\begin{minipage}[t]{1\columnwidth}%
\begin{tabular}{rll}
\texttt{the} & \texttt{dog chased} & ; 2 observations\tabularnewline
\texttt{the} & \texttt{dog killed } & ; 1 observation\tabularnewline
\end{tabular}%
\end{minipage} 

\medskip{}
and the 3-gram stalk for the cat is:

\medskip{}

\qquad{}%
\noindent\begin{minipage}[t]{1\columnwidth}%
\begin{tabular}{rll}
\texttt{chased the} & \texttt{cat} & ; 1 observation\tabularnewline
\texttt{the} & \texttt{cat chased } & ; 2 observations\tabularnewline
\texttt{the} & \texttt{cat killed } & ; 1 observation\tabularnewline
\end{tabular}%
\end{minipage} 

\medskip{}

These are now explicitly vectors, as the addition of the observation
count makes them so. The vertical alignment reminds us that they are
also still stalks, and that the vector comes from collocations. 

Recall how a vector is defined. One writes a vector $\vec{v}$ as
a sum over basis elements $\hat{e}_{i}$ with (usually real-number)
coefficients $a_{i}$:
\[
\vec{v}=\sum_{i}a_{i}\hat{e}_{i}
\]
The basis elements $\hat{e}_{i}$ are unit-length vectors. Another
common notation is the bra-ket notation, which says the same thing,
but in a different way:
\[
\vec{v}=\sum_{i}a_{i}\left|i\right\rangle 
\]
The bra-ket notation is slightly easier to use for this example. The
above 3-gram collocations can be written as vectors. The one for dog
would be 
\[
\overrightarrow{dog}=2\left|the\;*\;chased\right\rangle +\left|the\;*\;killed\right\rangle 
\]
while the one for cat would be

\[
\overrightarrow{cat}=\left|chased\;the\;*\right\rangle +2\left|the\;*\;chased\right\rangle +\left|the\;*\;killed\right\rangle 
\]
The $*$ here is the wild-card; it indicates where ``dog'' and ``cat''
should go, but it also indicates how the basis vectors should be treated:
the wild-card helps establish that dogs and cats are similar. It allows
the basis vectors to be explicitly compared to one-another. The ability
to compare these allows the dot product to be taken.

Recall the definition of a dot-product (the inner product). For $\vec{v}$
as above, and $\vec{w}=\sum_{i}b_{i}\hat{e}_{i}$, one has that
\[
\vec{v}\cdot\vec{w}=\sum_{i}\sum_{j}a_{i}b_{j}\hat{e}_{i}\cdot\hat{e_{j}}=\sum_{i}\sum_{j}a_{i}b_{j}\delta_{ij}=\sum_{i}a_{i}b_{i}
\]
where the Kronecker delta was used in the middle term:
\[
\hat{e}_{i}\cdot\hat{e_{j}}=\delta_{ij}=\begin{cases}
1 & \mbox{if }i=j\\
0 & \mbox{if }i\ne j
\end{cases}
\]
Thus, the inner product of $\overrightarrow{cat}$ and $\overrightarrow{dog}$
can be computed:
\[
\overrightarrow{cat}\cdot\overrightarrow{dog}=0\cdot1+2\cdot2+1\cdot1=5
\]
One common way to express the similarity of $\overrightarrow{cat}$
and $\overrightarrow{dog}$ is to compute the cosine similarity. The
angle $\theta$ between two vectors is given by
\[
\cos\theta=\vec{v}\cdot\vec{w}/\left|\vec{v}\right|\left|\vec{w}\right|
\]
where $\left|\vec{v}\right|=\sqrt{\sum_{i}a_{i}^{2}}$ is the length
of $\vec{v}$. Since $\left|\overrightarrow{cat}\right|=\sqrt{6}$
and $\left|\overrightarrow{dog}\right|=\sqrt{5}$ one finds that 
\[
\cos\theta=\frac{5}{\sqrt{30}}\approx0.913
\]
That is, dogs and cats really are similar. 

If one was working with a dependency parse, as opposed to 3-grams,
and if one used the Frobenius algebra notation such as that used by
Kartsaklis in \cite{Kart2014}, then one would write the basis elements
as a peculiar kind of tensor, and one might arrive at an expression
roughly of the form 
\[
\overline{dog}=2\left(\overleftarrow{the}\otimes\overrightarrow{chased}\right)+1\left(\overleftarrow{the}\otimes\overrightarrow{killed}\right)
\]
and 
\[
\overline{cat}=\left(\overleftarrow{chased}\otimes\overleftarrow{the}\right)+2\left(\overleftarrow{the}\otimes\overrightarrow{chased}\right)+1\left(\overleftarrow{the}\otimes\overrightarrow{killed}\right)
\]
Ignoring the differences in notation (ignoring that the quantities
in parenthesis are tensors), one clearly can see that these are still
feature vectors. Focusing on the vector aspect only, these represent
the same information as the 3-gram feature vectors. They're the same
thing. The dot products are the same, the vectors are the same. The
difference between them is that the bra-ket notation was used for
the 3-grams, while the tensor notation was used for the dependency
parse. The feature vectors can also be written using the link-grammar-inspired
notation: \\

\qquad{}%
\begin{minipage}[t]{0.8\columnwidth}%
\textsf{dog: {[}the- \& chased+{]}2 or {[}the- \& killed+{]}1;}

\textsf{cat: {[}chased- \& the-{]}1 or {[}the- \& chased+{]}2 or {[}the-
\& killed+{]}1;}%
\end{minipage}\\
\\
\\
The notation is different, but the meaning is the same. The above
gives two feature vectors, one for dog, and one for cat. They happen
to look identical to the 3-gram feature vectors because this example
was carefully arranged to allow this. In general, dependency parses
and 3-grams are going to be quite different; for these short phrases,
they happen to superficially look the same. In any of these cases,
and in any of these notations, the concept of feature vectors remain
the same.

\subsection*{Stalk fields and vector fields}

The figures \ref{fig:N-gram-base} and \ref{fig:Dependency-base}
illustrate the base space. Above each point in the base space, one
can, if one wishes, plant a stalk.

\begin{figure}[h]

\caption{Corn field; stalk field}

\includegraphics[width=0.4\columnwidth]{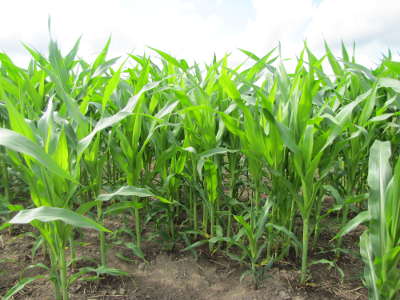} \qquad{}\includegraphics[width=0.25\columnwidth]{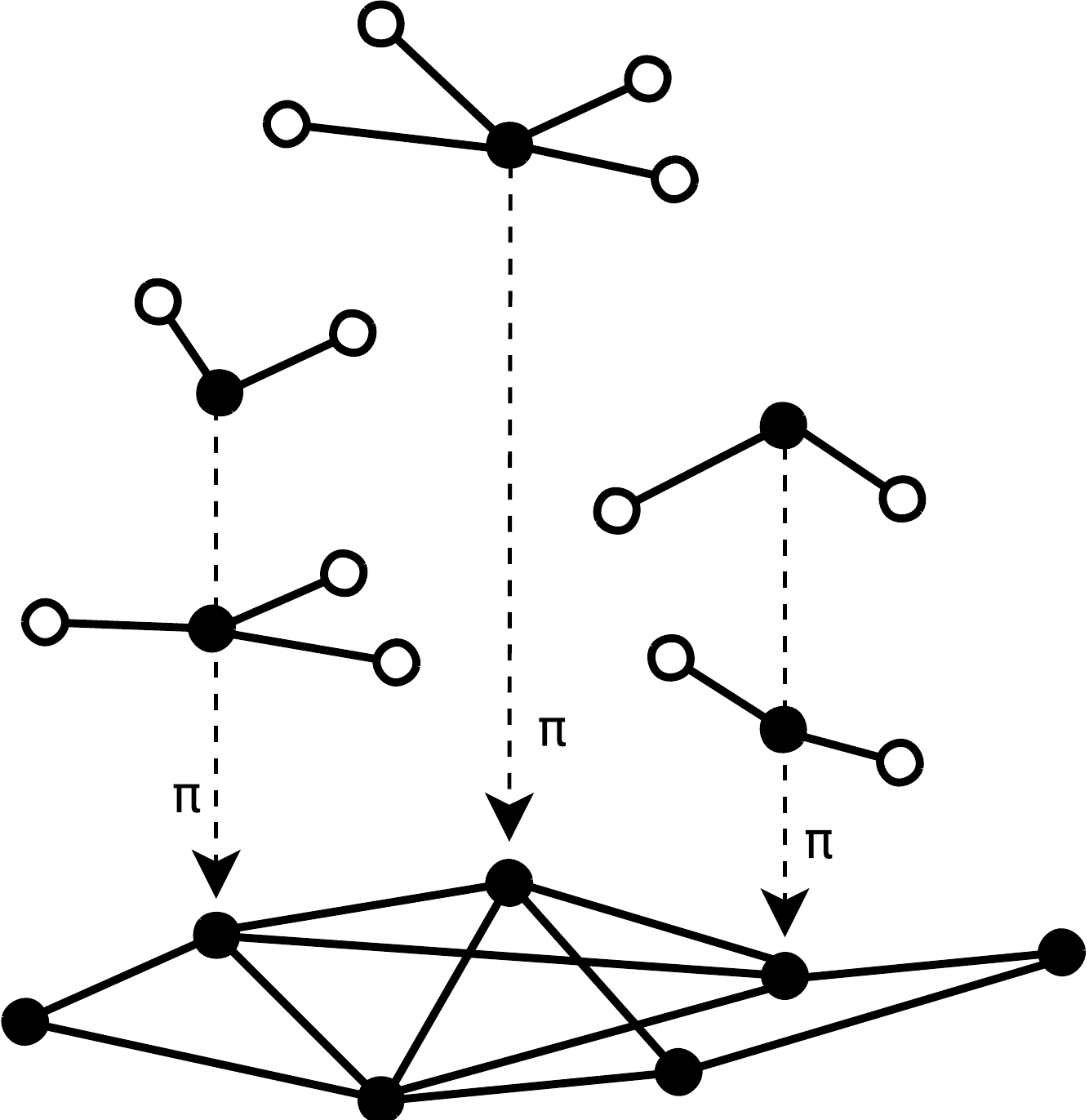}

\end{figure}

\begin{figure}[h]
\caption{Sheafs have big sections, in general\label{fig:General-Sheaf}}

\includegraphics[width=0.45\columnwidth]{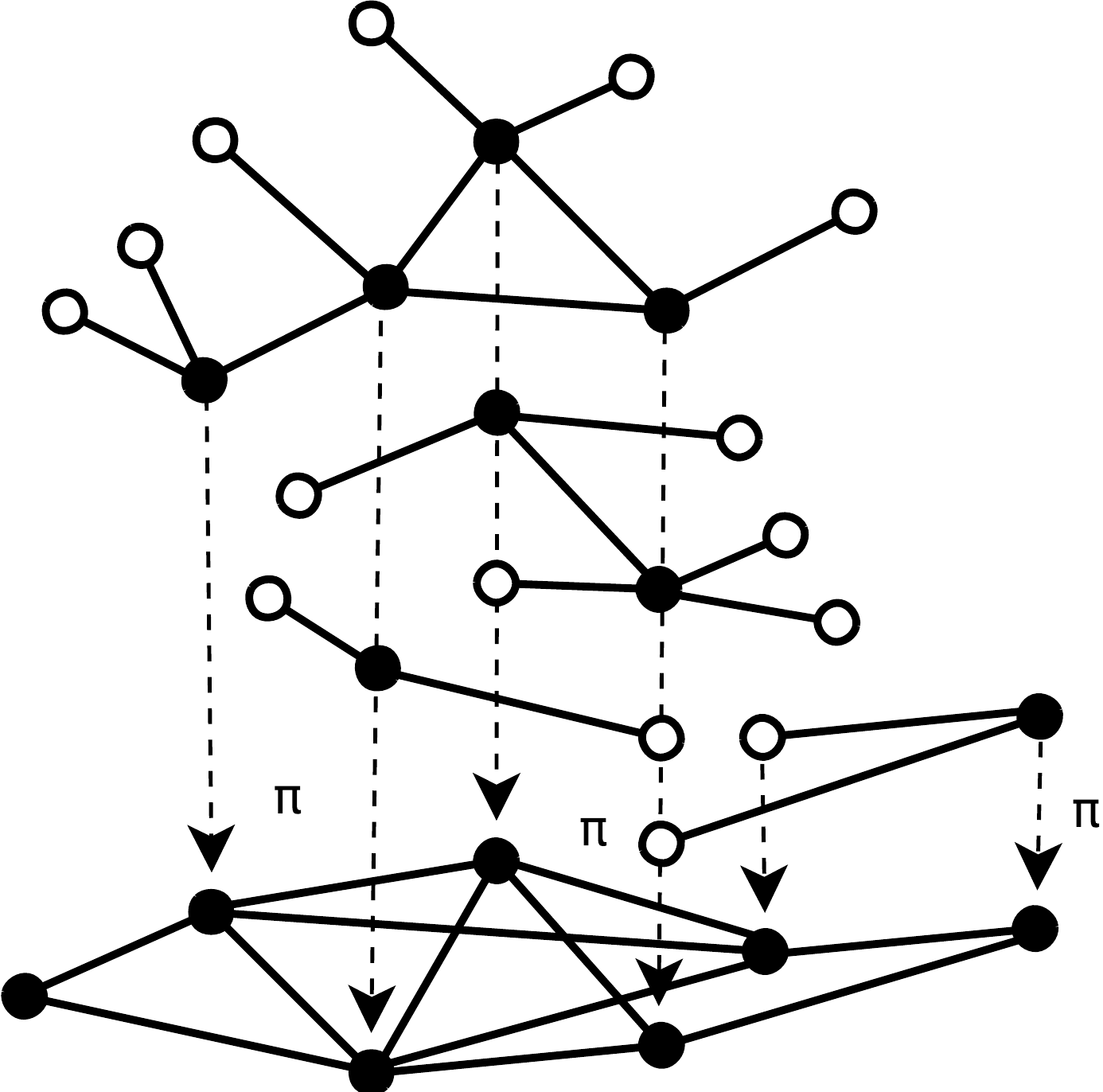}

\end{figure}

Such a plantation is not a sheaf; or rather it could be, but it is
not one with large sections. The stalk field only has individuals
seeds up and down each stalk; the stalks are not linked to one-another.
In the general case, illustrated in figure \ref{fig:General-Sheaf},
the stalks are linked to one-another; the sections really do start
to resemble sheets of paper stacked one on top another.

The general sheaf, as depicted here, holds much more data than just
the base space. It holds the data showing where the base space came
from: how the base space was a projection of sections. Holding such
a large amount of data might be impractical: in the previous example,
holding the parse data for 10 million individual, distinct sentences
might be a challenge. The stalk field is meant to be a half-way point:
it can hold more information than the base alone, but still be computationally
manageable. For example, the sme dataset discussed previously, containing
10 million sentences composed of 130K words has been found to contain
6 million seeds; these are observed on average of 2.5 times each,
although the distribution is roughly Zipfian: a few are observed hundreds
of thousands of times, and more than a third are observed only once. 

A particular appeal of the stalk field is that each stalk can be re-interpreted
as a vector. For each point of the base space, one just attaches a
single vector. There is no additional structure, and all this talk
of stalks can be brushed away as just a layer of theoretical complexity:
in the end, its just per-base-point feature vectors.

The power of the stalk representation is to keep in mind that the
basis elements are not just vacuous items, but are in fact jigsaw-puzzle
pieces that can be connected to one-another. Again, each stalk can
be viewed as a stack of jigsaw-puzzle pieces.

If there is a vector at each point, can the sheaf, as described here,
be thought of as a fiber bundle? Maybe, but that is not the intent.
In a fiber bundle, each fiber is isomorphic to every other. Thus,
locally, a fiber bundle always looks like the produce space $U\times F$
with $U\subseteq B$ and $F$ the fiber. Fiber bundles are interesting
when they are glued together in non-trivial ways, globally. Here,
there's a different set of concerns: its the local structure that
is interesting, and not so much the global structure. Also, there
has been no attempt to make each stalk (or stalk-space) isomorphic
to every other. If each stalk is a vector in a vector space, one could,
in principle, force that vector space to be the same, everywhere.
This does not buy much: in the practical case, the support for any
given vector is extremely sparse. 

In some cases, it is natural to have different stalks be incomparable.
In biology, some stalks may correspond to enzymes, others to RNA,
others to DNA. In some vague philosophical sense, it could be argued
that these are ``all the same'': examples of molecules. In practice,
forcing such unification seems to be a losing proposition. The goal
of the technology here is to detect, observe and model fine details
of structure, and not to mash everything into one bag.

\subsection*{Presheaves}

The formal definition of a sheaf entails a presentation of the so-called
``gluing axioms''. These are technical requirements that ensure
that the stalks can be linked, and sections projected in a ``common
sense'' kind of fashion. For example, if a section contains a sentence,
one expects that the sentence is grammatical. One also expects to
be able to extract phrases out of it. Gluing sentences together, one
expects to arrive at coherent paragraphs. In a biochemical setting,
one expects that all of the individual reactions in a pathway fit
together. One expects to be able to talk about subsets of the full
pathway without obtaining nonsense. This is just common sense.

Unfortunately, ``common sense'' being a commodity in short supply,
the gluing axioms must be written in detail. Before this can be done,
the axioms for a presheaf must be reviewed. There are several. Rather
than presenting these as axioms, they are presented below as ``claims''.
It is up to the reader to verify that the structures defined earlier
satisfy these claims. This is done for several reasons. First, such
proofs are a bit tedious, and would be out of place in this otherwise
rather informal treatment of the topic. Second, the overall informality
of this document gives little support for weighty proofs. Third, most
of these claims should be fairly self-evident, upon a bit of exploration.
Finally, many choices were left to the reader: should edges be directed?
Are they labeled? Do vertexes carry additional markings or values?
Each choice of labeling and marking potentially affects the verification
of these claims. Thus, the below are presented as ``claims'', living
in limbo between axioms and theorems.

First, a definition.
\begin{defn*}
An \noun{open subgraph} $U$ of a graph $G$ is defined to be a section
of $G$. $\diamond$
\end{defn*}
This definition helps avoid what would otherwise be confusing terminology.
The open subgraphs below will always be subgraphs of the base space
$B$. The open subgraphs are created by taking scissors and cutting
edges in the graph, but leaving the cut half-edges attached, as they
were originally. That is, the cut edges are converted into connectors.
By leaving these connectors in place, much of the information needed
to glue them back together remains intact. It is up to the reader
to convince themselves that these open subgraphs behave essentially
the same way as open sets in a topological space do: one can take
intersections and unions, and doing so still results in an open subgraph.
One can even build a Borel algebra out of them, but his will not be
needed.

The presheaf is defined in terms of a functor and it's properties. 
\begin{claim*}
There exists a functor $F$ such that, for each open subgraph $U$
of the base graph $B$, there exists some collection $F(U)$ of sections
above $U$. $\diamond$
\end{claim*}
Next, the restriction morphism, which cuts down or restricts this
collection.
\begin{claim*}
For each open subgraph $V\subseteq U$ there is a morphism $\mbox{res}_{V,U}:F\left(U\right)\to F\left(V\right)$.
$\diamond$
\end{claim*}
Since $V$ is smaller, we expect $F\left(V\right)$to be smaller,
also. The restriction morphism trims away the unwanted parts. The
trimming needs to stay faithful, to preserve the structure. Thus
\begin{claim*}
For every open subgraph $U$ of the base graph $B$, the restriction
morphism $\mbox{res}_{U,U}:F\left(U\right)\to F\left(U\right)$ is
the identity on $F\left(U\right)$. $\diamond$
\end{claim*}
The restrictions must compose in a natural way, as well, so that if
one trims a bit, then trims a bit more, its the same as doing it all
at once.
\begin{claim*}
For a sequence of open subgraphs $W\subseteq V\subseteq U$, the restrictions
compose so that $\mbox{res}_{W,V}\circ\mbox{res}_{V,U}=\mbox{res}_{W,U}$.
$\diamond$
\end{claim*}
If a system obeys the above, it is technically called a \noun{presheaf}.
A presheaf is much like the (informal) definition given for a sheaf,
above. However, it is possible to create structures that satisfy the
above claims (axioms), but don't quite match the intended definition
of a sheaf. In particular, the above are not enough to guarantee that
the sections in the presheaf can be organized properly into stalks.
To get well-behaved stalks, more is needed. These are the gluing axioms.

\subsection*{Gluing axioms}

The open subgraphs behave much like open sets. Thus, the concept of
an open covering can be imported in a straight-forward way. A collection
$\left\{ U_{i}\right\} $ of open subgraphs is an open cover for an
open subgraph $U$ if the union of all the $U_{i}$ contain $U$.
That is, they are an open cover if $U\subseteq\bigcup_{i}U_{i}$.
The union of open subgraphs is meant to be ``obvious'': join together
the connectors, where possible.

A presheaf is a sheaf if it obeys the following two claims/axioms.
\begin{claim*}
(Locality) If $\left\{ U_{i}\right\} $ is an open cover for $U$,
and if $s,t\in F\left(U\right)$ are sections such that $s\vert_{U_{i}}=t\vert_{U_{i}}$
for each $U_{i}$, then $s=t$. $\diamond$
\end{claim*}
In the above, the notation $s\vert_{V}$ denotes the restriction of
the section $s$ to the open subgraph $V$ of the base space $B$.
Pictorially, $s\vert_{V}$ is that part of the section that sits on
the stalks above $V$. It is a trimming-down of $s$ so that it projects
cleanly down to $V$ and to nothing larger. If each $U_{i}$ is a
seed in the base space, then $s\vert_{U_{i}}$ is a seed in the stalk
above $U_{i}$. Note that $s\vert_{U_{i}}$ might be the empty set.
The locality axiom is basically saying ``stalks exist''. Alternately,
the locality axiom says that if you cut up a layer-cake, you can still
tell, after the cutting, which layer was which.

The gluing axiom is needed to reassemble the pieces.
\begin{claim*}
(Gluing) If $\left\{ U_{i}\right\} $ is an open cover for $U$, and
if $s_{i}\in F\left(U_{i}\right)$ are sections restricted to each
$U_{i}$, and if, for all pairs $i,j$ the $s_{i}$ and $s_{j}$ agree
on overlaps, then there exists a section $s\in F\left(U\right)$ such
that $s_{i}=s\vert_{U_{i}}$. $\diamond$
\end{claim*}
In the above, the phrase ``$s_{i}$ and $s_{j}$ agree on overlaps''
means that $s_{i}\vert_{U_{i}\cap U_{j}}=s_{j}\vert_{U_{i}\cap U_{j}}$.
Note that $U_{i}\cap U_{j}$ might be the empty set, in which case
no agreement is needed. The gluing axioms states, more or less, that
if the layer cake is cut into pieces, and the pieces can be reassembled
with the edges lining up correctly, then the original layers can be
re-discovered.

Gluing is perhaps not as trivial as it sounds. It will be seen later
on that gluing is essentially the same thing as parsing. Obtaining
a successful parse is the same thing as assembling a valid section
ut of the parts. In the case of natural language, a parse succeeds
if and only if a sentence is grammatically valid. But of course! The
sections of a natural language sheaf are exactly the grammatical sentences.

Until this more detailed presentation of parsing is described, one
can imagine the following scenario. If seeds correspond to jigsaw-puzzle
pieces, then the sections $s_{i}$ correspond to partially-assembled
parts of the jigsaw. Two such parts $s_{i}$ and $s_{j}$ agree on
overlaps if $U_{i}\cap U_{j}$ is non-empty, and these two parts can
be joined together. If the connectors are typed, then there may be
multiple distinct connectors that can be joined to one-another. They
just might fit. That is, there might be more than one way to make
$s_{i}$ and $s_{j}$ connect, possibly by shifting, turning, the
pieces, etc. If one then tried to connect $s_{k}$, there might be
multiple ways of doing this, leading to a combinatorial explosion.
At some point in this process, one might discover that there is simply
no way at all to connect the next piece: it just won't fit. One then
has to back-track, and try a different arrangement. Obtaining an efficient
algorithm to perform this back-tracking is non-trivial: such algorithms
are called parsers, and gluing is parsing.

\subsection*{Does this really work?}

The sheaf axioms presented above are standardized and are presented
in many books. See, for example, Eisenbud \& Harris\cite{Eisenbud2000}
or Mac Lane \& Moerdijk\cite{MacLane1992}. The point of the above
is to convince the reader that the structures being described really
are sheaves, in the formal sense of the word. There's a big difference
though: everything above was developed from the point of view of graphs,
and that really does change the nature of the game. That said, the
reason that all of this machinery ``works'' is because the open
subgraphs really do behave very much like open sets. Because of this,
many concepts from topology extend naturally to the current structures.

This is not exactly a new realization. The ``open subgraphs'' defined
here essentially form a \href{https://en.wikipedia.org/wiki/Grothendieck_topology}{Grothendieck topology,}
and the thing that is being called a ``sheaf'' should probably be
more accurately called a ``site''. Developing and articulating this
further is left for a rainy day.

It is worth noting at this point that the normal notion of a ``germ''
in sheaf theory corresponds to what is called a ``seed'', here.
I suppose that the vocabulary used here could be changed, but I do
like thinking of seeds as sticky burrs. The biological germ of a seed
is that thing left, when the outer casing is removed.

The use of the jigsaw-puzzle piece analogy to define connectors is
strongly analogous to the construction of the \href{https://en.wikipedia.org/wiki/Nerve_of_a_covering}{Čech nerve}.
This can be thought of as a way of inducing overlaps from fiber products.
This point is returned to, later on.

\subsection*{Cohomology}

In orthodox mathematics, the only reason that sheaves are introduced
is to promptly usher the reader to Čech cohomology in the next chapter
of any book on algebraic topology. That won't be done here, so what's
the point of all this?

Well, this won't be done here mostly because I'm running out of space,
and, in the context of biology and linguistics, this is uncharted
territory. But some comments are in order. First, if the point of
this was merely to get at graph theory, there would not be much to
say. For example, the homotopy theory of graphs is more-or-less boring:
every graph is homotopic to a bouquet of circles. Homotopy and homology
on graphs only becomes interesting if one can add 2-cells and $n$-cells
for $n>1$; then one gets cellular homology. Can that ever happen
here?

If one considers biochemistry, and use the Krebs cycle (the citric
acid cycle) as an example, then the answer is yes. This is a loop;
it's essentially exothermic, or a kind of pump, in that the loop always
goes around in one direction. The edges are directional. Its a cycle
not only in a biological sense, but also in the mathematical sense:
it can be considered to be the boundary of a 2-cell. The Krebs cycle
is not the only cycle in biochemistry, and many of these cycles share
common edges. In essence, there's a whole bunch of 2-cells in biochemistry,
and they're all tangent to one-another. That is, there are chain complexes
in biochemistry. Is there interesting homology? Perhaps not, as this
would require some 2-cells to run ``backwards'', and that seems
unlikely. That would imply that there are no 3-cells in biochemistry.
But who knows; we have not had the tools to ``solve biochemistry''
before.

What about linguistics? Examples here seem to be more forced. Yes,
dependencies can be directional. Dependency trees are trees, however.
One can allow loops in them, but these loops are always acyclic. (\emph{viz.}
a ``DAG'' - a directed acyclic graph). There are no obviously cyclic
phenomena in natural language.

\subsection*{Why sheaves?}

By pointing out that natural language and biology can be described
with sheaves, it is hoped that this will prove better insights into
their structure, and provide a clear framework to think about the
structure of such data. 

For example, consider the normally vague idea of the ``language graph''.
What is this? One has dueling notions: the graph of all sentences;
the generative power of grammars. Sheaves provide a clearer picture:
the graph itself is the base space, while surface and deep structure
can be explored through sections.

It can be argued that orthodox corpus linguistics studies the sheaf
of surface structure, with especially strong focus on the stalks.
Differences in the stalks reveal differences between regional dialects.
Much more interesting is that the corpus linguists have analyzed stalks
to discover not just differences in socio-economic status, but even
to discover politically-motivated speech, truth and lack-thereof in
journalism and news media.\cite{Louw2007}

The orthodox corpus linguists are not interested in refining their
collocations into a generative grammar. One does not obtain a generative
model of how different speakers in different socio-economic classes
speak; corpus linguistics examples are just that: examples that are
not further refined. By applying a pattern mining approach, the underlying
grammar can be discovered computationally. By viewing structure holistically,
as a sheaf, one can see ways in which this might be done. 

Besides the sheaf of surface realizations studied by corpus linguists,
there are several different kinds of sheaves of grammatical structure.
Each section is a grammatically valid sentence, expressed as a tree
or as a DAG (directed acyclic graph) of some sort, annotated with
additional information, based on the formalities of that particular
grammatical approach (dependency grammar, head-phrase-structure grammar,
etc). The orthodox approach is to view the grammar as being the primary
object of study. The sheaf approach helps emphasize how that grammar
was arrived at: distinct words were grouped into grammatical classes.
Put differently, distinct stalks are recognized as being very similar,
if not identical, and are merged together to form a grammatical category;
it is no longer individual words that link with one-another, but the
grammatical classes.

Viewing language as a sheaf helps identify how one can automatically
extract grammatical classes: If one can judge two stalks as being
sufficiently similar in some way, then one can merge them into one,
proceeding in this way to create a reduced, concentrated model of
language that captures it's syntactic structure.

One can do even more: one can play off the differences in regional
dialects, or differences due to social-economic classes, discovered
by statistical means from a corpus, and attach these to specific grammatical
structures, identified from syntactic analysis. That is, by seeing
both activities: surface realizations and deeper structure as two
slightly different forms of ``the same thing'', one can see-saw,
lever ones way about, moving from one to the other and back. Tools
can be developed that do both, instead of just one or just the other.
One can actually unify into one, what seem to be very theories and
approaches, and one can develop the techniques to move between these
theories. This seems to be a very big win.

\section*{clustering morphisms}

The primary topic of this part is that the extraction of structure
from data is more-or-less a kind of morphism between sheaves. A ``pseudo-morphism''
might be an more appropriate term, as the definition here will not
be axiomatically precise.

There are several types of morphisms that are of interest. one kind
keeps the base space intact, but attempts to map one kind of section
into another: for example, mapping sections of $n$-grams into sections
of dependency parses. This resembles the orthodox concept of a morphism
between sheaves. The other kind of morphism is one that attempts to
re-arrange the base space, by grouping together multiple stalks into
one. This second kind of morphism is the one discussed in this part.
It is roughly termed a ``clustering morphism''.

There are several kinds of clustering morphisms that are interesting.
One was previously illustrated. Starting with

\[
\overrightarrow{Mary}\otimes\overline{walked}\otimes\overrightarrow{home}+\overrightarrow{Mary}\otimes\overline{ran}\otimes\overrightarrow{home}+\overrightarrow{Mary}\otimes\overline{drove}\otimes\overrightarrow{home}
\]
one wishes to deduce 

\[
\overrightarrow{Mary}\otimes\left(\overline{walked}+\overline{ran}+\overline{drove}\right)\otimes\overrightarrow{home}
\]
This seems to be relatively straight-forward to accomplish, as it
looks like a simple application of the distributive law of multiplication
over addition. It is perhaps deceptive, as it presumes that the three
words ``walked'', ``ran'', ``drove'' do not appear anywhere
else in the sheaf.

A different example is that of forcing diagonalization where there
is none. Given a structure such as
\[
\left|Mary\right\rangle \otimes\left|walked\right\rangle +\left|Adam\right\rangle \otimes\left|ran\right\rangle 
\]
one wishes to induce
\[
\left(\left|Mary\right\rangle +\left|Adam\right\rangle \right)\otimes\left(\left|walked\right\rangle +\left|ran\right\rangle \right)
\]
This resembles a grammar-school error: an inappropriate application
of the distributive law. But is it really? Part of the problem here
is that the notation itself is biased: the symbols $\otimes$ and
$+$ look like the symbols for multiplication and addition, and we
are deeply ingrained, from childhood - from grammar school, that multiplication
distributes over addition, but not the other way around. By using
these symbols, one introduces a prejudice into one's thinking; the
prejudice suggests that one operation is manifestly legal, while the
other is dubious and requires lots of justification.

This prejudice can run very deeply: in data-mining software, if not
in the theories themselves, the first operation might be hard-coded
into the software, into the theory, and assumed to be \emph{de facto}
correct. By contrast, the second relation seems to require data-mining,
and maybe lots of it: crunching immense, untold numbers of examples
to arrive at the conclusion that such a diagonalization is valid.
Perhaps reality is somewhere between these two extremes: the first
factorization should not be assumed, and, as a result, the second
diagonalization might not be so hard to discover. Perhaps induction
can be applied uniformly to both cases.

\subsection*{Induction}

The goal of machine learning in data science is the induction of the
factorization and diagonalization from a given dataset. Both examples
given above are misleading, because they ignore the fact that they
are embedded in a much larger corpus of language. How might these
two cases be induced from first principles, \emph{ab initio}, from
nothing at all, except for a bunch of examples? 

On possibility is to start by looking for pair-wise correlations.
This works: that is how $\left|Mary\right\rangle \otimes\left|walked\right\rangle $
is discovered in the first place: these two words were collocated.
Likewise, for $\left|Adam\right\rangle \otimes\left|ran\right\rangle $.
But what about inducing diagonalization? Here, one observes that Mary
does lots of things, and so does Adam. Writing down the collocation
stalk for Mary, and the one for Adam should indicate that these two
stalks are quite similar. How can similarity be judged? The cosine
distance, previously reviewed, is a plausible way to start. One can
legitimately conclude that Adam and Mary belong in the same grammatical
category. What about ``walked'' and ``ran''? One can create a
stalk for these two as well, and it should not be hard to conclude,
using either cosine distance, or something else, that the two are
quite similar.

Great. Now what? Just because Adam and Mary are similar, and ``ran''
and ``walked'' are similar, this is still not quite enough to justify
the diagonalization. After all, ``Mary ran'' and ``Adam walked''
have not been observed; how can one justify that these will likely
be observed, which is the central claim that diagonalization is making?

The answer would need to be that certain cross-correlations are only
weakly seen. Define the set of named-things, and action-things, already
discovered: $names=\{Adam,Mary\}$ while $actions=\{ran,walked\}$.
Let the $\lnot$ symbol denote ``not'', so that $\lnot names$ is
the set of all things are not $names$, and $\lnot actions$ denote
all things that are not actions. Consider then the correlation matrix

\medskip{}

\begin{center}
\begin{tabular}{c|c|c|}
\multicolumn{1}{c}{} & \multicolumn{1}{c}{$actions$} & \multicolumn{1}{c}{$\lnot actions$ }\tabularnewline
\cline{2-3} 
$names$ & High & Low\tabularnewline
\cline{2-3} 
 $\lnot names$ & Low & n/a\tabularnewline
\cline{2-3} 
\end{tabular}
\par\end{center}

\medskip{}
The entry ``High'' means that a large amount of correlation is observed,
while ``Low'' means that little is observed. Correlation can be
measured in many ways; mutual information and Kullbeck-Liebler divergence
are popular. 

Why might this work? The point is that if $\lnot actions$ contains
words like $book$ or $tree$, then sentences like ``Mary book''
or ``Adam tree'' are not likely to be observed; if $\lnot names$
includes words like $green$ or $the$, then sentences like ``green
walked'' or ``the ran'' should be rare. 

The correlation matrix embodies the very meaning of ``diagonalization'':
a matrix is diagonal, when the entries along the diagonal are large,
and the entries not on the diagonal are zero. Observing this structure
then justifies writing $names\otimes actions$, which is exactly what
one wanted to induce. Can one also validly claim that $\left(\lnot names\right)\otimes\left(\lnot actions\right)$?
Well, probably not. The correlation there might be low - pairs would
be inconsistent as to how compatible they are. It might be hard to
compute, and, in the current context, it seems not to be wanted.

Can one induce factorization in the same way? Factorization, as given
above, seemed ``obvious'', but that was only due to the use of symbols
that prejudiced one's thinking. Factorization is, in fact, every bit
as non-obvious as diagonalization. The reason it seems so obvious
in the example was that the corpus ``Mary walked home'', etc. did
not include any sentences about Adam, nor anything about ``to the
store'', ``to work'', etc. Once these are included, factorization
starts to look a lot like diagonalization, if not exactly the same
thing. Inducing a subject-verb-object relationship can be done by
means of correlation, but is harder to depict, because the correlation
is no longer a pair-wise matrix, but is 3D, forming a cube, because
three categories need to be compared: $names$, $actions$, and $places$,
where $places=\{home,to\,the\,store,to\,work\}$. This is shown below.

\medskip{}

\begin{flushleft}
\qquad{}$places\;\begin{cases}
\\
\\
\\
\end{cases}$  %
\begin{tabular}{c|c|c|}
\multicolumn{1}{c}{} & \multicolumn{1}{c}{$actions$} & \multicolumn{1}{c}{$\lnot actions$ }\tabularnewline
\cline{2-3} 
$names$ & High & Low\tabularnewline
\cline{2-3} 
 $\lnot names$ & Low & n/a\tabularnewline
\cline{2-3} 
\end{tabular}
\par\end{flushleft}

\medskip{}

\medskip{}

\begin{flushleft}
\qquad{}\qquad{}\qquad{}\qquad{}\qquad{}$\lnot places\;\begin{cases}
\\
\\
\\
\end{cases}$ %
\begin{tabular}{c|c|c|}
\multicolumn{1}{c}{} & \multicolumn{1}{c}{$actions$} & \multicolumn{1}{c}{$\lnot actions$ }\tabularnewline
\cline{2-3} 
$names$ & Low & n/a\tabularnewline
\cline{2-3} 
 $\lnot names$ & n/a & n/a\tabularnewline
\cline{2-3} 
\end{tabular}
\par\end{flushleft}

\medskip{}
That is, one can induce a three-way relationship $(x,y,z)$ whenever
that relationship is frequently seen, and all three of the relations
$(\lnot x,y,z)$, $(x,\lnot y,z)$ and $(x,y,\lnot z)$ are not seen.
This extends to 4-way relations, and so on.

There is one notable phenomenon that is not covered by the above:
words that have different meanings, but the same spelling, for example,
``saw'' or ``fly'' which are both nouns and verbs. This complicates
the approach above; this issue is returned to in a later section,
titled \nameref{sec:Polymorphism}.

\subsection*{Related concept: Discrimination}

Several comments are in order. The above presents grammatical induction
as a form of discrimination - \href{https://en.wikipedia.org/wiki/Binary_classification}{binary discrimination},
even, which is considered to be a particularly simple form of learning.
There are many available techniques for this, and one can promptly
fall into the examination of \href{https://en.wikipedia.org/wiki/Receiver_operating_characteristic}{ROC curves},
and the like. It is important to note that what is being sketched
here is the idea of discrimination in the context of sheaves, and
not the idea of binary discrimination as some panacea for linguistics.

The above was also vague as to the form of correlation: how should
it be done? Should it literally be correlation, in the sense of probability
theory? Should it be mutual information? Something else? This is left
intentionally vague: different measures of correlation are possible.
Some may produce better results than others. A general theoretical
framework is being sketched here; the quality of different algorithms
is not assessed or presented. It is up to the reader to experiment
with different forms of correlation and discrimination.

\subsection*{Related concept: Clustering}

The induction, described above, resembles the machine-learning concept
of clustering in several ways. There are also some strong differences,
and so this is worth reviewing. Two old and time-honored approaches
to clustering are support vector machines (SVM) and $k$-means clustering.
The first relies explicitly on some sort of vectorial representation
for the data, while the second expects some sort of metric for judging
whether two points are similar or not. For the former, interpreting
the stalks as the feature vectors is sufficient, while for the latter,
the cosine distance can fill the role of a metric.

These two approaches are sufficient to extract classes of things,
such as $names$, $places$ and $actions$ in the above example. The
accuracy of the extracted categories is rarely excellent, but is certainly
adequate enough to proceed to other stages. Except ... that's it.
These clustering techniques stop there; they say nothing at all about
inducing grammatical relations. To induce grammatical relations, one
\emph{also} has to perform discrimination in some way. One has to
combine the results obtained from clustering, and then discriminate
to induce grammar.

Note that the discrimination step provides information about how good
the clustering was. Say, for example, that cosine distance was used,
together with $k$-means clustering, to obtain classes of words. Was
this clustering ``adequate''? That question can be answered by examining
the ROC curves obtained from a binary discrimination step. Different
kinds of clustering will present different ROC curves. This can be
used as feedback for the clustering step, so that one gets a recursive
learning step, alternating between discrimination and clustering.

This observation of recursion, of course, raises the question: can
clustering and discrimination be combined into one effective algorithm?
Yes, they can.

\subsection*{Related concepts: neurlal nets, adagram.}

Besides binary discrimination, there are other approaches. Approaches
that are more sophisticated include decision trees and decision forests.
These two approaches treat the vectors as tables of input data, and
then pick and choose among the vector components deemed predictive.

x

x

foo

orig neural net: \cite{Bengio2003}

\subsection*{More}

This: https://becominghuman.ai/how-does-word2vecs-skip-gram-work-f92e0525def4

\subsection*{Why clustering?}

foo-bar

x

x

By contrast, the goal here is not just to talk about a graph $G$
relative to a single $A$, but relative to a huge number of different
$A$'s. What's more, the internal structure of these $A$'s will continue
to be interesting, and so is carried onwards. Finally, the act of
merging together multiple vertexes into one $A$ may result in some
of the existing edges being cut, or new edges being created. The clustering
operation applied to the graph alters the graph structure. These considerations
are what makes it convenient to abandon traditional graph theory,
and to replace it by the notion of sheaves and sections.

x

The above establishes a vocabulary, a means for talking about the
clustering of similar things on graphs. It does not suggest how to
cluster. Without this vocabulary, it can be very confusing to visualize
and talk about what is meant by clustering on a graph. Its worth reviewing
some examples.
\begin{itemize}
\item In a social graph, a cluster might be a clique of friends. By placing
these friends into one group, the stalk allows you to examine how
different groups interact with one-another.
\item In proteomic or genomic data, if one can group together similar proteins
or genes into clusters, one can accomplish a form of dimensional reduction,
simplifying the network model of the dataset. It provides a way to
formalize network construction, without the bad smell of ad-hoc simplifications.
\item In linguistic data, the natural clustering is that of words that behave
in a similar syntactic fashion; such clusters are commonly called
``grammatical classes'' or ``parts of speech''. In particular,
it allows one to visualize language as a graph. So: consider, for
example, the set of all dependency parses of all sentences in some
corpus, say Wikipedia. Each dependency parse is a tree; the vertexes
are words, and the edges are the dependencies. Taken as a graph, this
is a huge graph, with words connecting to other words, all over the
place. Its not terribly interesting in this raw state, because its
overwhelmingly large. However, we might notice that all sentences
containing the word ``dish'' resemble all sentences containing the
word ``plate''; that these two words always get used in a similar
or the same way. Grouping these two words together into one reduces
the size of the graph by one vertex. Aggressively merging similar
words together can sharply shrink the size of the graph to a manageable
size. One gets something more: the resulting graph can be understood
as encapsulating the structure of the English language. 
\end{itemize}
This last example is worth expanding on. Two things happen when the
compressed graph is created. First, that graph encodes the syntactic
structure of the language: the links between grammatical classes indicate
how words can be arranged into grammatically correct sentences. Second,
the amount of compression applied can reveal different kinds of structures.
With extremely heavy compression, one might discover only the crudest
parts of speech: determiners, adjectives, nouns, transitive and intransitive
verbs. Each of these classes are distinct, because they link differently.
However, if instead, a lot less compression is applied, then one can
discover synonymous words: so, ``plate'' and ``dish'' might be
grouped together, possibly with ``saucer'', but not with ``cup''.
Here, one is extracting a semantic grouping, rather than a syntactic
grouping. 

So, the answer to ``why clustering?'' is that it allows information
to be extracted from a graph, and encoded in a useful, usable fashion.
No attempt is made here to suggest how to cluster; merely, that if
an equivalence relation is available, and if it is employed wisely,
then one can construct quotient graphs that encode important relationships
of the original, raw graph.

\section*{Types}

It is notationally awkward to have to write stalks in terms of the
sets of vertexes that they are composed of; it is convenient to instead
replace each set by a symbol. The symbol will be called a \noun{type}.
As it happens, these types can be seen to be the same things occurring
in the study of type theory; the name is justified.

The core idea can be illustrated with Link Grammar as an example.
The Link Grammar disjuncts \emph{are} one and the same thing as stalks.
It is worth making this very explicit. A subset of the Link Grammar
English dictionary looks like this:\medskip{}

\begin{minipage}[t]{0.5\columnwidth}%
\textsf{cat dog: D- \& S+;}

\textsf{the a: D+;}

\textsf{ran: S-;}%
\end{minipage}\medskip{}
\\
This states that ``cat'' and ``dog'' are both vertexes, and they
are in the same stalk. That stalk has two connectors: \textsf{D-}
and \textsf{S+}, which encode the other stalks that can be connected
to. So, the \textsf{D+} can be connected to the \textsf{D-} to form
a link. The link has the form \textsf{(\{the, a\}, \{cat, dog\})}
and the connector symbols \textsf{D+} and \textsf{D-} act as abbreviations
for the vertex sets that the unconnected end can connect to. The +
and - symbols indicate a directionality: to the right or to the left.
The capture the notion that, in English, the word-order matters. To
properly explain the + and -, we should have to go back to the definition
of a graph on the very first page, and introduce the notion of left-right
order among the vertices. Doing so from the very beginning would do
nothing but clutter up the presentation, so that is not done. The
reader is now invited to treat the initial definition of the graph
as a monad: there are additional details ``under the covers'', but
they are wrapped up and ignored, and only the relevant bits are exposed.
Perhaps the vertices had a color. Perhaps they had a name, or a numerical
weight; this is ignored. Here, we unwrap the idea that the vertices
must be organized in a left-right order. Its sufficient, for now,
to leave it at that.
\begin{figure}[h]
\caption{Three stalks and two typed links}
\includegraphics[width=0.4\columnwidth]{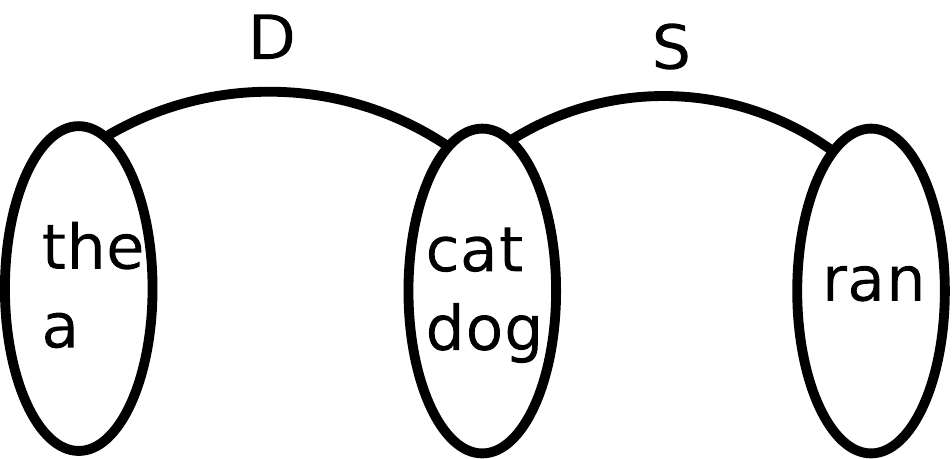}

\end{figure}

The three stalks here encode a set of grammatically valid English
language sentences. Hooking together the S- and S+ connectors to form
an S link, one obtains the sequence \textsf{{[}\{the, a\} \{cat, dog\}
\{ran\}{]}}. This can be used to generate grammatically valid sentences:
pick one word from each set, and one gets a valid sentence. Alternatively,
this structure can be taken to encode the sum-total knowledge about
this toy language: it is a kind-of graphical representation of the
entire language, viewed as a whole.
\begin{defn*}
Given a stalk $S=\left(V,L\right)$, the \noun{connector type} of
$L$ is a symbol that can be used as a synonym for the set $L$. It
serves as a short-hand notation for $L$ itself. $\diamond$
\end{defn*}
Just as in type theory, a type can be viewed a set. Yet, just as in
type theory, this is the wrong viewpoint: a type is better understood
as expressing a property: it is an intensional, rather than an extensional
description. Formally, in the case of finite sets, this may feel like
splitting hairs. For an intuitive understanding, however, its useful
to think of a type as a property carried by an object, not just the
class that the object can be assigned to.

\subsection*{Why types? }

Types are introduced here primarily as a convenience for working with
stalks. They are labels, but they can be useful. Re-examining the
examples:
\begin{itemize}
\item In a social graph, one group of friends might be called ``students''
and another group of friends might be called ``teachers''. The class
labels are useful for noting the function and relationship of the
different social groups.
\item In a genetic regulatory network, sub-networks can be classified as
\textquotedbl{}positive regulatory pathways\textquotedbl{} or \textquotedbl{}negative
regulatory pathways\textquotedbl{} with respect to the activation
of a particular gene.
\end{itemize}
These examples suggest that the use of types is little more than a
convenient labeling system. In fact, more hay can be made here, as
types interact strongly with category theory: types are used to describe
the internal language of monoidal categories. But this is a rather
abstract viewpoint, of no immediate short-term use. Suffice it to
say that appearance of types in grammatical analysis of a language
is not accidental. 

\subsection*{What kind of information do types carry?}

The above example oversimplifies the notion of types, presenting them
as a purely syntactic device. In practice, types also carry semantic
information. The amount of semantic information varies inversely to
the broadness of the type. In language, coarse-grained types (noun,
verb) carry almost no semantic information. Fine-grained types carry
much more: a ``transitive verb taking a particle and an indirect
object'' is quite specific: it must be some action that can be performed
on some object using some tool in some fashion. An example would be
``John sang a song to Mary on his guitar'': there is a what, who
and and how yoked together in the verb ``sang''. The more fine-grained
the classification, the more semantic content is contained in it.

This suggests that the proper approach is hierarchical: a fine-grained
clustering, that captures semantic content, followed by a coarser
clustering that erases much of this, leaving behind only ``syntactic''
content.

\section*{Parsing}

The introduction remarked that not every collection of seeds can be
assembled in such a way as to create a valid graph. This idea can
be firmed up, and defined more carefully. Generically, a valid assembly
of seeds is called a parse, and the act of assembling them is called
parsing, which is done by parse algorithms. To illustrate the process,
consider the following two seeds:\\

\qquad{}%
\begin{minipage}[t]{0.8\columnwidth}%
$v_{2}:\left\{ \left(v_{2},v_{1}\right),\left(v_{2},v_{3}\right)\right\} $

$v_{3}:\left\{ \left(v_{3},v_{2}\right)\right\} $%
\end{minipage}\\
\\
Represented graphically, these seeds are

\begin{figure}[H]
\caption{Two unconnected seeds}
\includegraphics[width=0.55\columnwidth]{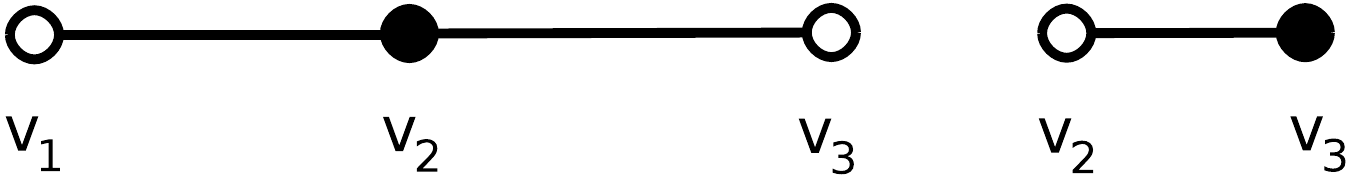}

\end{figure}
The connector (half-edge) $\left(v_{2},v_{3}\right)$ appears with
both polarities, and can be linked together to form a link. The connector
$\left(v_{2},v_{1}\right)$ has nothing to connect to. Even after
maximally linking these two seeds, one does not obtain a valid graph:
the vertex $v_{1}$ is missing from the vertex-set of the graph, even
though there is an edge ready to attach to it. This provides an example
of a failed parse. It is enough to add the seed $v_{1}:\left\{ \left(v_{1},v_{2}\right)\right\} $
to convert this into a successful parse. Adding this seed, and then
attempting to maximally link it results in a valid graph; the parse
is successful.

\begin{figure}[h]
\caption{Parsing is the creation of links}
\includegraphics[width=0.45\columnwidth]{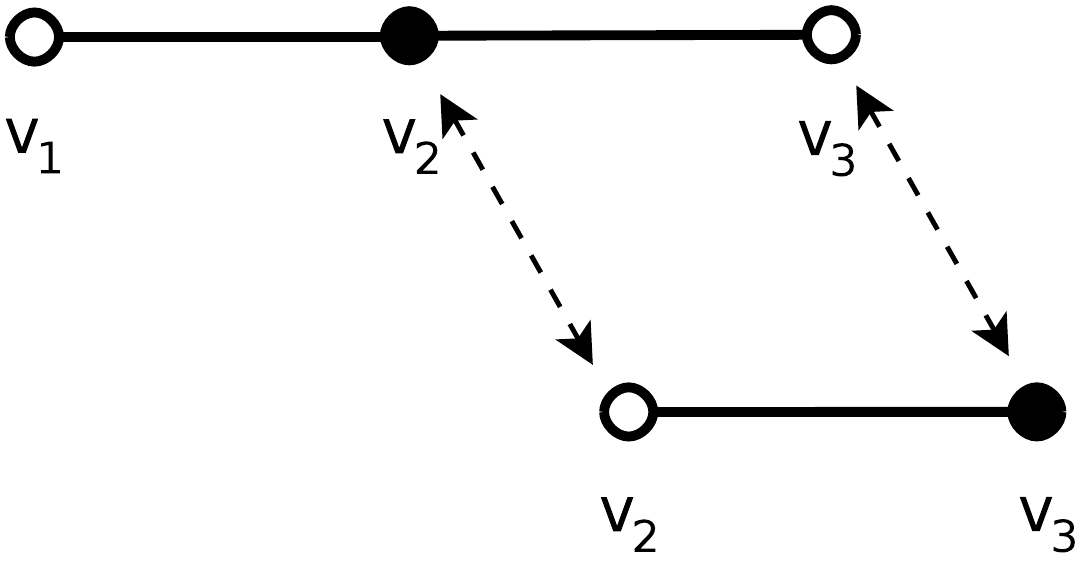}

\end{figure}

Note the minor change in notation: the colon is used as a separator,
with the germ appearing on the left, and set of connectors on the
right. The relevance of this notational change becomes more apparent,
if we label the vertexes in a funny way: let $v_{1}$ carry the label
``the'', and $v_{2}$ carry the label ``dog'' and $v_{3}$ carry
the label ``ran''. The failed parse is meant to illustrate that
``dog ran'' is not a grammatically valid sentence, whereas ``the
dog ran'' is. 

Converting these seeds to also enforce left-right word-order requires
the notation\\

\qquad{}%
\begin{minipage}[t]{0.8\columnwidth}%
\textsf{the: \{(the, dog+)\}}

\textsf{dog: \{(dog, the-), (dog,ran+)\}}

\textsf{ran: \{(ran, dog-)\}}%
\end{minipage}\\
\\

This notation is verbose, and slightly confusing. Repeating the germ
as the first vertex in every connector is entirely unnecessary. Write
instead:\\

\qquad{}%
\begin{minipage}[t]{0.8\columnwidth}%
\textsf{the: \{ dog+ \}}

\textsf{dog: \{ the-, ran+\}}

\textsf{ran: \{ dog- \}}%
\end{minipage}\\
\\

The set-builder notation is unneeded, and perhaps slightly confusing.
In particular, the word ``dog'' has two connectors on it; both must
be connected to obtain a valid parse. The ampersand can be used to
indicate the requirement that both connectors are required. This notation
will also be useful in the next section.\\

\qquad{}%
\begin{minipage}[t]{0.8\columnwidth}%
\textsf{the: dog+ ;}

\textsf{dog: the- \& ran+ ;}

\textsf{ran: dog- ;}%
\end{minipage}\\
\\

This brings us almost back to the previous section, but not quite.
Here, we are working with seeds; previously we worked with stalks.
Here, the connector type labels were not employed. In real-world use-cases,
using stalks and type labels is much more convenient.

This now brings us to a first draft of a parse algorithm. Given an
input set of vertices, it attempts to find a graph that is able to
connect all of them.
\begin{enumerate}
\item Provide a dictionary $D$ consisting of a set of unconnected stalks. 
\item Input a set of vertices $V=\left\{ v_{1},v_{2},\cdots,v_{k}\right\} $. 
\item For each vertex in $V$, locate a stalk which contains that vertex
in it's germ.
\item Attempt to connect all connectors in the selected stalks.
\item If all connectors can be connected, the parse is successful; else
the parse fails.
\item Print the resulting graph. This graph can be described as a pair $\left(V,E\right)$
with $V$ the input set of vertexes, and $E$ the set of links obtained
from fully connecting the selected stalks.
\end{enumerate}
The above algorithm is ``generic'', and does not suggest any optimal
strategy for the crucial steps 3 or 4. It also omits discussion of
any further constraints that might need to be applied: perhaps the
edges need to be directed; perhaps the resulting graph must be a planar
graph (no intersecting edges); perhaps the graph must be a minimum
spanning tree; perhaps the input vertexes must be arranged in linear
order. These are additional constraints that will typically be required
in some specific application.

\subsection*{Why parsing?}

The benefit of parsing for the analysis of the structure of natural
language is well established. Thus, an example of parsing in a non-linguistic
domain is useful. Consider having used the above graph compression/vertex-edge
clustering techniques to obtain a collection of stalks that describe
genomic interactions. This collection provides the initial dictionary
$D$. Now imagine a process where a certain specific set of genes
are associated with some particular trait or reaction. Is this a complete
set? Can it be said that their interactions are fully understood?

One way to answer these last two questions would be to apply the parse
algorithm, using the known dictionary, to see if a complete interaction
network can be obtained. If so, then this new specific gene-set fits
the general pattern. If not, if a complete parse cannot be found,
then one strongly suspects that there remain one or more genes, yet
undetermined, that also play a role in the trait. To find these, one
might examine the stalks that might have been required to complete
the parse: these will give hints as to the specific type of gene,
or style of interaction to search for.

Thus, parsing new gene expressions and pathways offers a way of discovering
whether they resemble existing, known pathways, or whether they are
truly novel. If they seem novel, parsing also gives strong hints as
to where to look for any missing pieces or interactions.

\subsection*{Is this really parsing?}

The above description of parsing is sufficiently different from standard
textbook expositions of natural language parsing that some form of
an apology needs to be written. 

The first step is to observe that the presented algorithm is essentially
a simplified, generalized variation of the Link Grammar parsing algorithm.\cite{Sleator1993}
The generalization consists in the removal of word-order and link-crossing
constraints.

The second step is to observe that the theory of Link Grammar is more-or-less
isomorphic to the theory of pregroup grammars\cite{Kart2014} (see
\href{https://en.wikipedia.org/wiki/Pregroup_grammar}{Wikipedia});
the primary differences being notational. The left-right directional
Link Grammar connectors correspond to the left and right adjoints
in a pregroup. A Link Grammar disjunct (that is, a seed) corresponds
to a sequence of types in a pregroup grammar. The correspondence is
more-or-less direct, except that link grammar is notationally simpler
to work with.

The third step is to observe that the Link Grammar is a form of dependency
grammar. Although the original Link Grammar formulation uses undirected
links, it is straight-forward and unambiguous to mark up the links
with head-dependent directional arrows.

The fourth step is to realize that dependency grammars (DG) and head-phrase-structure
grammars (HPSG) are essentially isomorphic. Given one, one can obtain
the other in a purely mechanistic way. 

The final step is to realize that most introductory textbooks describe
parsers for a context-free grammar, and that, for general instructional
purposes, such parsers are sufficient to work with HPSG. The priamry
issue with HPSG and context-free language parsers is that they obscure
the notion of linking together pieces; this is one reason why dependency
grammars are often favored: they make clear that it is the linkage
between various words that has a primary psychological role in the
human understanding of language. It should be noted that many researchers
in the psychology of linguistics are particularly drawn to the categorial
grammars; these are quite similar to the pregroup grammars, and are
more closely related to Link Grammar than to the phrase-structure
grammars.

\section*{Polymorphism\label{sec:Polymorphism}}

Any given vertex may participate in two or more seeds, independently
from one-another. It is this statement that further sharpens the departure
from naive graph theory. This is best illustrated by a practical example.

\begin{figure}
\caption{Polymorphism\label{fig:Polymorphism}}

\includegraphics[width=0.75\columnwidth]{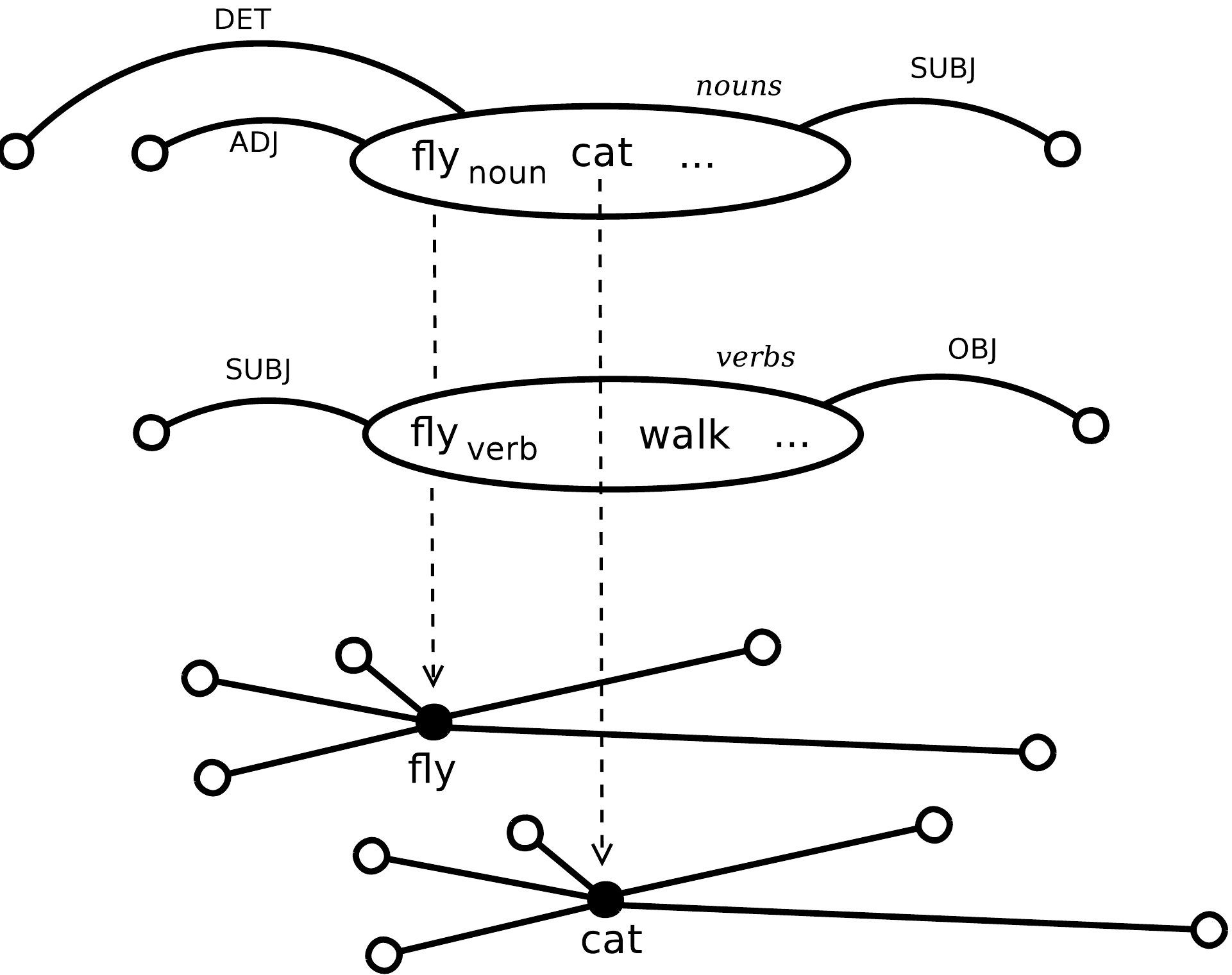}

This figure illustrates a polymorphic assignment for the word ``fly''.
It is split into two parts, the first, a noun, classed with other
nouns, showing labeled connectors to determiners, adjectives, and
a connector showing that nouns can act as the subject of a verb. The
second class shows labeled connectors to subjects and objects, as
is appropriate for transitive verbs. Underneath are the flattened
raw seeds, showing the words ``fly'' and ``cat'' and the myriad
of connectors on them. The flattened seeds cannot lead to grammatical
linkages, as they mash together into one the connectors for different
parts of speech.

\rule[0.5ex]{1\columnwidth}{1pt}

\end{figure}

Consider a large graph, constructed from a large corpus of English
language sentences. As subgraphs, it might contain the two sentences
``A big fly landed on his nose'' and ``It will fly home''. The
vertex ``fly'' occurs as a noun (the subject, with dterminer and
adjective) in one sentence, and a verb (with subject and object) in
the other. Suppose that the equivalence relation, described in the
clustering section, also has the power to discern that this one word
should really be split into two, namely $fly_{\mbox{noun}}$ and $fly_{\mbox{verb}}$,
and placed into two different stalks, namely, in the ``noun'' stalk
in the first case, and the ``verb'' stalk in the second. Recall
that these two stalks must be different, because the kinds of connectors
that are allowed on a noun must necessarily be quite different from
those on a verb. One is then lead to the image shown in figure \ref{fig:Polymorphism}. 

The point of the figure is to illustrate that, although the ``base
graph'' may not distinguish one variant of a vertex from another,
it is important to discover, extract and represent this difference.
The concept of ``polymorphism'' applies, because the base vertex
behaves as one of several distinct types in practice. There are several
ways the above diagram can be represented textually. As before, the
Link Grammar-style notation is used, as it is fairly clear and direct.
One representation would be to expose the polymorphism only in the
connectors, and not in the base vertex label:

\medskip{}

\qquad{}%
\noindent\begin{minipage}[t]{1\columnwidth}%
\texttt{fly: (DET- \& ADJ- \& SUBJ+) or (SUBJ- \& OBJ+);}%
\end{minipage} 

\medskip{}

A different possibility is to promptly split the vertex label into
two, and ignore the subscript during the parsing stage:

\medskip{}

\qquad{}%
\noindent\begin{minipage}[t]{1\columnwidth}%
\texttt{fly.noun cat: (DET- \& ADJ- \& SUBJ+);}

\texttt{fly.verb walk: (SUBJ- \& OBJ+);}%
\end{minipage} 

\medskip{}

Either way, the non-subscripted version of $fly$ behaves in a polymorphic
fashion.

Note that the use of the notation ``\texttt{or}'' to disjoin the
possibilities denotes a choice function, and not a boolean-or. That
is, one can choose either one form, or the other; one cannot choose
both. During the parse, both possibilities need to be considered,
but only one selected in the end. This implies that at least some
fragment of linear logic is at play, and not boolean logic. (this
should be expanded upon in future drafts).

\subsection*{Similar concept: part of speech}

It is tempting to identify the connectors DET, ADJ, SUBJ, OBJ in the
diagrams above with ``parts of speech''. This would be a mistake.
In conventional grammatical analysis, there are half-a-dozen or a
dozen parts of speech that are recognized: noun, verb, adjective,
and so on. By contrast, these connector types indicate a grammatical
role. That is, the disjunct \texttt{SUBJ- \& OBJ+} indicates a word
that takes both a subject and an object: a transitive verb. That is,
the disjunct is in essence a fine-grained part of speech, indicating
not only verb-ness, but the specific type of verb-ness (transitive).

The Link Grammar English dictionary documents more than 100 connector
types, these are subtyped, so that approximately 500 connectors might
be seen. These connectors, when arranged into disjuncts, result in
tens of thousands of disjuncts. That is, Link Grammar defines tens
of thousands of distinct ``parts of speech''. The can be thought
of as parts of speech, but they are quite fine-grained, far more fine-grained
than any text on grammar might ever care to list.

If one uses a technique, such as MST parsing\cite{Yuret1998}, and
then extracts disjuncts, one might observe more than 6 million disjuncts
and 9 million seeds on a vocabulary of 140K words. These are, again,
in the above technical sense, just ``parts of speech'', but they
are hyperfine-grained. The count is overwhelming. So, although it
is techinically correct to call them ``parts of speech'', it is
a conceptual error to think of a class that has six million representatives
as if it were a class with a dozen members.

\subsection*{Similar concept: skip-grams}

The N-gram\cite{Rosen1996} and the more efficient skip-gram\cite{Guthrie2006}
models of semantic analysis provide somewhat similar tools for understanding
connectivity, and differentiating different forms of connectivity.
In a skip-gram model, one might extract two skip-grams from the above
example sentences: ``a fly landed'' and ``it fly home''. A clustering
process, such as adagram or word2vec might be used to classify these
two strings into distinct clusters, categorizing one with other noun-like
words, and the other with verb-like words. 

The N-gram or skip-gram technique works only for linear, sequenced
data, which is sufficient for natural language, but cannot be employed
in a generic non-ordered graphical setting. To make this clear: a
seed representation for the above would be: ``fly: a- landed+''
indicating that the word ``a'' (written as the connector ``a-'')
comes sequentially before ``fly'', while the word ``landed'' (written
as the connector ``landed+'') comes after. The other phrase has
the representation ``fly: it- home+''. These two can now be employed
in a clustering algorithm, to determine whether they fall into the
same, or into different categories. If one treats the skip-grams,
and the seeds as merely two different representations of the same
data, then applying the same algorithm to either should give essentially
the same results.

The seed representation, however, is superior in two different ways.
First, it can be used for non-sequential data. Second, by making clear
the relationship between the vertex and its connectors, the connectors
can be treated as ``additional data'', tagging the vertex, carrying
additional bits of information. That additional information is manifested
from the overall graph structure, and is explicit. By contrast, untagged
N-grams or untagged skip-grams leave all such structure implicit and
hidden.

\subsection*{Polymorphism and semantics}

The concept of polymorphism introduced above lays a foundation for
semantics, for extracting meaning from graphs. This is already hinted
at by the fact that any English-language dictionary will provide at
least two different definitions for ``fly'': one tagged as a noun,
the other as a verb. The observation of hyperfine-grained parts of
speech can push this agressively farther. 

In a modern corpus of English, one might expect to observe the seeds
``apple: green-'' and ``apple: iphone+''. The disjuncts ``green-''
and ``iphone+'' can be interpreted as a kind-of tag on the word
``apple''. Since there are exactly two tags in this example, they
can be viewed as supplying exactly one bit of additional information
to the word ``apple''. Effectively, a single apple has been split
into two distinct apples. Are they really distinct, however? This
can only be judged on the basis of some clustering algorithm that
can assign tagged words to classes. Even very naive, unsophisticated
algorithms might be expected to classify these two different kinds
of apple into different classes; the extra bit of information carried
by the disjunct is a bit of actual, usable information.

To summarize: the arrangement of vertexes into polymorphic seeds and
sections enables the vertexes to be tagged with extra information.
The tags are the connectors themselves: thier presence or absence
carries information. That extra information can be treated as ``semantic
information'', identifying different types or kinds, rather than
as purely syntactic information about arrangments and relationships.

\section*{Conclusion}

This document presents a way of thinking about graphs that allows
them to be decomposed into constituent parts fairly easily, and then
brought together and reassembled in a coherent, syntactically correct
fashion. It does so without having to play favorites among competing
algorithmic approaches and scoring functions. It makes only one base
assumption: that knowledge can be extracted at a symbolic level from
pair-wise relationships between events or objects.

It touches briefly, all too briefly, on several closely-related topics,
such as the application of category theory and type theory to the
analysis of graph structure. These topics could be greatly expanded
upon, possibly clarifying much of this content. It is now known to
category theorists that there is a close relationship between categories,
the internal languages that they encode, and that these are reflections
of one another, reflecting through a theory of types. A reasonable
but incomplete reference for some of this material is the HoTT book.
It exposes types in greater detail, but does not cover the relationship
between internal languages, parsing, and the modal logic descriptions
of parsing. It is possible that there are texts in proof theory that
cover these topics, but I am not aware of any.

This is a bit unfortunate, since I feel that much or most of what
is written here is ``well known'' to computational proof theorists;
unfortunately, that literature is not aimed at the data-mining and
machine-learning crowd that this document tries to address. Additions,
corrections and revisions are welcomed.

\bibliographystyle{tufte}

\end{document}